\documentclass[10pt,twocolumn,letterpaper]{article}

\usepackage{wacv}
\usepackage{times}
\usepackage{epsfig}
\usepackage{graphicx}
\usepackage{amsmath}
\usepackage{amssymb}
\usepackage{algorithm}
\usepackage{algpseudocode}
\usepackage{enumitem}

\usepackage[accsupp]{axessibility}

\usepackage[font=small,labelfont=small,skip=0pt,labelsep=period]{caption}
\usepackage{subcaption}

\usepackage{multirow}
\usepackage{booktabs}
\usepackage{tabularx}
\usepackage{array}

\newcolumntype{P}[1]{>{\centering\arraybackslash}p{#1}}

\usepackage[pagebackref=true,breaklinks=true,letterpaper=true,colorlinks,bookmarks=false]{hyperref}
\makeatletter
\@namedef{ver@everyshi.sty}{}
\makeatother
\usepackage{tikz,pgfplots}
\usetikzlibrary{plotmarks,shapes,arrows,positioning,3d}
\usetikzlibrary{arrows.meta}
\usetikzlibrary{decorations.text}
\pgfplotsset{compat=newest}

\usepackage{bm}
\newcommand{\mathbold}[1]{\bm{#1}}
\newcommand{\mbf}[1]{\mathbf{#1}}
\newcommand{\vect}[1]{\mbf{#1}}
\newcommand{\vectb}[1]{\mathbold{#1}}

\newcommand{\T}{\top}

\newcommand{\R}{\mathbb{R}}

\DeclareMathOperator{\diag}{diag}

\newcommand{\vgamma}[0]{\mathbold{\gamma}}

\newcommand{\vepsilon}[0]{\mathbold{\varepsilon}}

\newcommand{\vomega}[0]{\mathbold{\omega}}

\newcommand{\vpi}[0]{\mathbold{\pi}}

\newcommand{\va}{\mbf{a}}
\newcommand{\vb}{\mbf{b}}

\newcommand{\vf}{\mbf{f}}

\newcommand{\vh}{\mbf{h}}

\newcommand{\vm}{\mbf{m}}

\newcommand{\vp}{\mbf{p}}
\newcommand{\vq}{\mbf{q}}
\newcommand{\vr}{\mbf{r}}

\newcommand{\vu}{\mbf{u}}
\newcommand{\vv}{\mbf{v}}

\newcommand{\vx}{\mbf{x}}
\newcommand{\vy}{\mbf{y}}

\newcommand{\MA}{\mbf{A}}

\newcommand{\MI}{\mbf{I}}

\newcommand{\MM}{\mbf{M}}
\newcommand{\MP}{\mbf{P}}

\newcommand{\MR}{\mbf{R}}

\newcommand{\MT}{\mbf{T}}

\usepackage{xspace}
\renewcommand{\cf}{\textit{cf.}\xspace}

\usepackage{color}
\newcommand{\todo}[1]{\textcolor{blue}{\bf TODO: #1}\PackageWarning{TODO:}{#1!}}

\newlength\figureheight
\newlength\figurewidth

\renewcommand{\paragraph}[1]{\medskip\noindent{\bf#1}\ \ }

\usepackage[capitalize,nameinlink]{cleveref}
\crefname{section}{Sec.}{Sec.}
\crefname{proposition}{Prop.}{Props.}
\crefname{lemma}{Lem.}{Lems.}
\crefname{model}{Mod.}{Mods.}
\crefname{appendix}{App.}{Apps.}
\crefname{algorithm}{Alg.}{Algs.}

\renewcommand{\ref}[1]{\todo{Use cref instead of ref.}}
\renewcommand{\eqref}[1]{\todo{Use cref instead of eqref.}}

\def\epssuf{-eps-converted-to}

\def\wacvPaperID{276}

\wacvfinalcopy
\wacvfinalcopy

\ifwacvfinal

\fi

\ifwacvfinal
\usepackage[breaklinks=true,bookmarks=false]{hyperref}
\else
\usepackage[pagebackref=true,breaklinks=true,colorlinks,bookmarks=false]{hyperref}
\fi

\ifwacvfinal
\else
\fi

\begin{document}

\def\ourmethodname{HybVIO}
\def\thetitle{\ourmethodname: Pushing the Limits of Real-time Visual-inertial Odometry}
\title{\thetitle}
\ifwacvfinal

\author{Otto Seiskari$^1$ ~~ Pekka Rantalankila$^1$ ~~ Juho Kannala$^{1,2}$ ~~ Jerry Ylilammi$^1$ ~~ Esa Rahtu$^{1,3}$ ~~ Arno Solin$^{1,2}$ \\[6pt]
\begin{minipage}{.3\textwidth}\centering
$^1$Spectacular AI \\
{\footnotesize\tt \{first.last\}@spectacularai.com}
\end{minipage}
\hfill
\begin{minipage}{.3\textwidth}\centering
$^2$Aalto University \\
{\footnotesize\tt \{first.last\}@aalto.fi}
\end{minipage}
\hfill
\begin{minipage}{.3\textwidth}\centering
$^3$Tampere University \\
{\footnotesize\tt \footnotesize esa.rahtu@tuni.fi}
\end{minipage}
}

\else
\author{Anonymous}
\fi

\maketitle

\ifwacvfinal
\thispagestyle{empty}
\fi

\begin{abstract}
We present \ourmethodname, a novel hybrid approach for combining filtering-based visual-inertial odometry (VIO) with optimization-based SLAM. The core of our method is highly robust, independent VIO with improved IMU bias modeling, outlier rejection, stationarity detection, and feature track selection, which is adjustable to run on embedded hardware. Long-term consistency is achieved with a loosely-coupled SLAM module.
In academic benchmarks, our solution yields excellent performance in all categories, especially in the real-time use case, where we outperform the current state-of-the-art.
We also demonstrate the feasibility of VIO for vehicular tracking on consumer-grade hardware using a custom dataset, and show good performance in comparison to current commercial VISLAM alternatives.
\end{abstract}

\vspace*{-1em}

\section{Introduction}
\label{sec:intro}
\noindent
\emph{Visual-inertial odometry} (VIO) refers to the tracking of the position and orientation of a device using one or more cameras and an \emph{inertial measurement unit} (IMU), which, in this context, is assumed to comprise of at least an accelerometer and a gyroscope. A closely related term is \emph{visual-inertial SLAM} (VISLAM), which is typically used to describe methods that possess a longer memory than VIO: simultaneously with tracking, they produce a map of the environment, which can be used to correct accumulated drift in the case the device revisits a previously mapped area. Without additional inputs, these methods can only estimate the location relative to the starting point but provide no global position information. In the visual--inertial context, the orientation of the device also has one unsolvable degree of freedom: the rotation about the gravity axis, or equivalently, the initial compass heading of the device.

VISLAM is the basic building block of infrastructure-free augmented reality applications.
VIO, especially when fused with satellite navigation (GNSS), can be applied to tracking of various types of both industrial and personal vehicles, where it can maintain accurate tracking during GNSS outages, for instance, in highway tunnels. One of the main advantages of VIO over pure inertial navigation (INS), and consequently, the advantage of GNSS-VIO over GNSS-INS, is improved long-range accuracy. VIO can provide similar accuracy with consumer-grade hardware, than INS with high-end IMUs that are prohibitively expensive for any consumer applications.

\begin{figure}[t]
  \hspace{2em}
  \includegraphics[width=0.7\columnwidth]{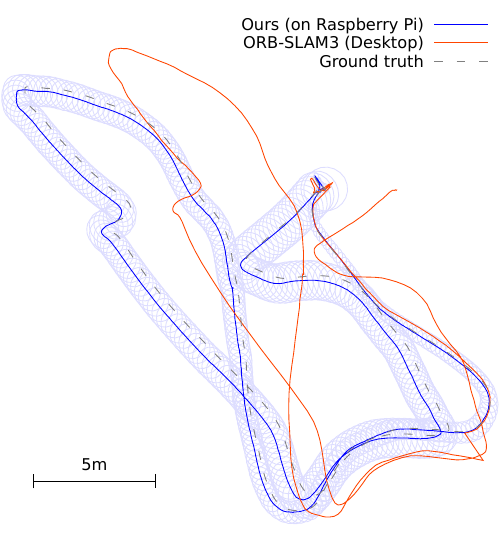}
  \vspace{-4pt}
  \caption{Real-time VIO with uncertainty quantification on an embedded processor compared to ORB-SLAM3 on a desktop CPU. The latter remains the leader in the EuRoC post-processing category, while our method yields the best online results.}
  \label{fig:teaser-euroc-uncertainty}
  \vspace*{-1em}
\end{figure}

The contributions of this paper are as follows.
  {\em (i)}~We extend the probabilistic inertial-visual odometry (PIVO) methodology  from monocular-only to stereo.
  {\em (ii)}~We improve the IMU bias modeling in PIVO with Ornstein-Uhlenbeck random walk processes.
  {\em (iii)}~We derive improved mechanisms for outlier detection, stationarity detection, and feature track selection that leverage the unique features of the probabilistic framework.
  {\em (iv)}~We present a novel hybrid method for ego-motion estimation, where extended Kalman filtering based VIO is combined with optimization-based SLAM.

These methods enable state-of-the-art performance in various use cases (online, offline, monocular, and stereo).
In particular, we outperform the previous leader, BASALT, in EuRoC MAV.
We also demonstrate the vehicular tracking capabilities of our VIO module with consumer-grade hardware, as well as our accuracy compared to commercial alternatives, using a custom dataset.

\section{Related work}\label{sec:related}\noindent
Our VIO module is a stereoscopic extension of PIVO~\cite{PIVO} and, consequently, a member of the MSCKF family of VIO methods that stems from~\cite{MSCKF}. Other recent methods belonging to the same class include the hybrid-EKF-SLAM (\cf~\cite{hybrid-ekf-slam}) method LARVIO~\cite{LARVIO} and S-MSCKF~\cite{stereoMSCKF}, which extends the original MSCKF to stereo cameras. These methods, following their EKF-SLAM predecessors (\eg,~\cite{monoSLAM}), use an Extended Kalman Filter (EKF) to keep track of the VIO state. They track the Bayesian conditional mean (CM) of the VIO state and keep it in memory together with its full covariance matrix, which limits the practical dimension of the state vector in real-time use cases.

An alternative to the above \emph{filtering-based} methods are \emph{optimization-based} approaches, which compute a Maximum A Posteriori (MAP) estimate in place of the conditional mean, and instead of storing a full covariance matrix, they may use sparse Bayesian factor graphs. The optimization-based methods are often stated to be more accurate than filtering-based methods and many recent publications prefer this approach. Notable examples include OKVIS~\cite{OKVIS}, MARS-VINS~\cite{MARS-VINS}, ORB-SLAM3~\cite{ORB-SLAM3}, BASALT~\cite{BASALT} and Kimera-VIO~\cite{Kimera}.

However, there are also disadvantages compared to filtering-based methods, namely, the lack of uncertainty quantification capabilities and the difficulty of marginalizing the active state on all the past data. Our method includes elements from both approaches, as filtering-based VIO is loosely coupled with optimization-based SLAM module. Previously, good results for post-processed trajectories have been reported with hybrid filtering--optimization approaches in, \eg,~\cite{Quan-et-al-2019} and \cite{Bai-et-al-2019}. However, our approach differs from these tightly-coupled solutions.
In addition to the more common \emph{sparse} approaches above, various alternatives have been proposed (see, \eg, \cite{ROVIO, vinet-2017, LSD-SLAM, Tanskanen-2015, VI-DSO}). For a more extensive survey of recent and historical methods, we refer the reader to \cite{ORB-SLAM3} and the references therein.

\section{Method description}
\label{sec:methods}\noindent

\subsection{VIO state definition}
\label{sec:vio-state}\noindent
Similarly to~\cite{PIVO} and~\cite{MSCKF}, we construct the VIO state vector at time step $t_k$,
\begin{equation}
\label{eq:vio-state}
  \vx_k = (\vpi^{(0)}_k, \vv_k, \vb_k, \tau_k, \vpi^{(1)}_k, \ldots, \vpi^{(n_a)}_k),
\end{equation}
using the poses $\vpi^{(j)}_k = (\vp^{(j)}_k, \vq^{(j)}_k) \in \R^3\times\R^4$ of the IMU sensor at the latest input sample ($j=0$) and a fixed-size window of recent camera frames ($j=1,\ldots,n_a$). The other elements in \cref{eq:vio-state} are the current velocity $\vv_k \in \R^3$,
a vector of IMU biases $\vb_k = (\vb^{\rm a}_k, \vb^{\omega}_k, \diag(\mbf{T}^{\rm a}_k))$
(see \cref{eq:imu-propagation}),
and an IMU-camera time shift parameter $\tau_k \in \R$, utilized as described in~\cite{imu-camera-time-shift-paper}.

In the EKF framework, the probability distribution of the state, given all the observations $\vy_{1:k}$ until time $t_k$, is modeled as Gaussian, $\vx_k|\vy_{1:k} \sim \mathcal N (\vm_{k|k}, \MP_{k|k})$. We model the orientation quaternions as Gaussians in $\R^4$ and restore their unit length after each EKF update step.

\subsection{IMU propagation model}
\label{sec:imu-propagation}
\noindent%
The VIO system is initialized to $(\vm_{1|1}, \MP_{1|1})$, where the current orientation $\vq^{(0)}_{1|1}$ is based on the first IMU samples equally to~\cite{INS-paper}. The other components of $\vm_{1|1}$ are fixed (zero or one), and $\MP_{1|1}$ is a fixed diagonal matrix.
No other measures are required to initialize the system.

Following~\cite{PIVO}, IMU propagation is performed on each synchronized pair $(\vomega_k, \va_k)$ of gyroscope and accelerometer samples as an EKF prediction step of the form
\begin{equation}
  \vx_{k|k-1} = \vf_k(\vx_{k-1|k-1}, \vepsilon_k)
\end{equation}
with $\vepsilon_k \sim \mathcal N(\mbf{0}, \mbf Q \Delta t_k)$.
The function $\mbf{f}_k$ updates the pose and velocity by the mechanization equation
\begin{equation}
  \label{eq:imu-propagation}
  \begin{pmatrix}
    \vect{p}_k \\ \vect{v}_k \\ \vect{q}_k
  \end{pmatrix}
  =
  \begin{pmatrix}
    \vect{p}_{k-1} + \vect{v}_{k-1}\Delta t_k \\
    \vect{v}_{k-1} + [\vect{q}_k (\tilde{\vect{a}}_k + \vectb{\varepsilon}^\mathrm{a}_k) \vect{q}_k^\star - \vect{g}] \Delta t_k \\
    \vectb{\Omega}[(\tilde{\vectb{\omega}}_k + \vectb{\varepsilon}^\omega_k) \Delta t_k] \vect{q}_{k-1}
  \end{pmatrix},
\end{equation}
where the bias-corrected IMU measurements are computed as $\tilde{\vect{a}}_k = \vect{T}_k^\mathrm{a} \, \vect{a}_k - \vect{b}^\mathrm{a}_k$ and $\tilde{\vectb{\omega}}_k = \vectb{\omega}_k - \vect{b}^\omega_k$.
In our model, the multiplicative correction $\vect{T}_k^\mathrm{a} \in \R^{3\times3}$ is a diagonal matrix.

Contrary to the approach used in~\cite{MSCKF}, this does not involve linearization errors that could cause the orientation quaternion to lose its unit length, since $\vectb{\Omega}[\cdot] \in \R^{4 \times 4}$ (\cf~\cite{Titterton+Weston:2004} or \cref{eq:quaternion-rotated-by-vector}) is an orthogonal matrix.

As an extension to~\cite{PIVO}, we assume the following model for the IMU biases:
\begin{equation}
  \label{eq:imu-bias-propagation}
  \begin{pmatrix}
    \vect{b}^\mathrm{a}_k \\ \vect{b}^\omega_k \\ \vect{T}_k^\mathrm{a}
  \end{pmatrix}
  =
  \begin{pmatrix}
    \exp(-\alpha_{\rm a}\Delta t_k) \vect{b}^\mathrm{a}_{k-1} + \epsilon_k^{\rm a} \\
    \exp(-\alpha_{\omega}\Delta t_k) \vect{b}^\omega_{k-1} + \epsilon_k^{\omega} \\
    \vect{T}_{k-1}^\mathrm{a}
  \end{pmatrix},
\end{equation}
where the parameters $\alpha, \sigma$ in the Ornstein--Uhlenbeck~\cite{Ornstein-Uhlenbeck-1930} random walks $\epsilon_k \sim \mathcal N(\mathbf{0}, \frac{\sigma^2}{2\alpha}[1 - \exp(-2\alpha\Delta t_k)])$ can be adjusted to match the characteristics of the IMU sensor.

\subsection{Feature tracking}
\label{sec:feature-tracking}
\noindent
Similarly to~\cite{MSCKF}, our visual updates are based on the constraints induced by viewing certain point features from multiple camera frames. We first utilize the \emph{Good Features to Track} (GFTT) algorithm~\cite{Shi+Tomasi:1994} (or, alternatively, FAST~\cite{FAST-features}, \cf\ \cref{tbl:parameters}) for detecting an initial set of features, which are subsequently tracked between consecutive frames using the pyramidal Lucas--Kanade method~\cite{Lucas+Kanade:1981} as implemented in the OpenCV library~\cite{OpenCV}. We also use the reprojections of previously triangulated 3D positions (\cf \cref{sec:visual-updates}) of the features as initial values for the LK tracker whenever available, which improves its accuracy and robustness, especially during rapid camera movements.

As in~\cite{PIVO}, features lost due to falling out of the view of the camera, or any other reason, are replaced by detecting new key points whenever the number of tracked features falls below a certain threshold. A minimum distance between features is also imposed in the detection phase and sub-pixel adjustment is performed on the new features.

In the case of stereo data, we detect the new features in the left camera frame, and find the matching points in the right camera frame, also using the Lucas--Kanade algorithm. This technique allows the use of raw camera images without a separate stereo rectification phase. The temporal tracking is only performed on the left camera frames, and the matches in the right frame are recomputed on each image. In addition, we reject features with incorrect stereo matches based on an epipolar constraint check.

Unlike~\cite{PIVO}, we additionally utilize a 3-point stereo RANSAC method described in~\cite{Nister+Naroditsky+Bergen:2004} or, in the monocular case, a mixture of 2-point (rotation only, \cf\ \cite{Kanatani-rotation-fitting-ransac2}) and 5-point RANSAC methods \cite{Nister}, for rejecting outlier features.

\subsection{Visual update track selection}
\label{sec:visual-update-track-selection}
\noindent
A feature track $\vy^j$ with index $j \in \mathbb N$ is, in the monocular case, a list of pixel coordinates ($\vy^j_i \in \R^2$), or pairs of coordinates $\vy^{j,R}_i, \vy^{j,L}_i \in \R^2$ in stereo.
The track is valid for a range of camera frame indices $i = i_{\min}^j, \ldots, i_{\max}^j$, where $i_{\min}^j$ corresponds to the frame where the feature is first detected and $i_{\max}^j$ the last frame where it is successfully tracked. In the stereo case, both $\vy^{j,R}$ and $\vy^{j,L}$ must be continuously tracked as described in \cref{sec:feature-tracking}.

On camera frame $i \leq i_{\max}^j$, denote by $b(i)$ the camera frame index of the last pose $\vpi^{(n_a)}_{k(i)}$ stored in the VIO state, and by $b(i,j) = \max(b(i), i_{\min}^j)$ the corresponding minimum valid camera frame index for track $j$.
Unlike~\cite{PIVO}, we do not always use all key points in the track, but select the subset of indices
\begin{equation}
  \label{eq:anti-track-reuse-criterion}
  S(i, j) = \{ b(i, j) \} \cup \{ \max(S(i', j)) + 1, \ldots, i \},
\end{equation}
where $i' < i$ is the last frame on which feature $j$ was used for a visual update (see \cref{sec:visual-updates}). In other words, we avoid ``re-using" the parts of the visual tracks that have already been fused to the filter state in previous visual updates.

Instead of using all available tracks on frame $i$ (denoted here by $U_i$), we pick the indices at random from the subset
\begin{equation}
  \label{eq:longer-than-median-tracks}
  \{ j \in U_i | L(i, j) > \mbox{median}_{U_i}(L(i, \cdot)) \},
\end{equation}
corresponding to longer-than-median tracks, where the length metric is defined as
\begin{equation}\textstyle
  \label{eq:track-length-metric}
  L(i, j) = \sum_{l \in S(i, j) \setminus \{ b(i, j) \}  } \| \vy^j_l - \vy^j_{l-1} \|_1.
\end{equation}
In the case of stereo data, \cref{eq:track-length-metric} is computed from the left camera features only ($\vy^j_l := \vy^{j,L}_l$).
Tracks are picked until the target number of visual updates have succeeded or the maximum number of attempts has been reached. Contrary to~\cite{MSCKF}, our visual update is performed individually on each selected feature track.

\begin{figure}
  \centering
  \includegraphics[width=\linewidth]{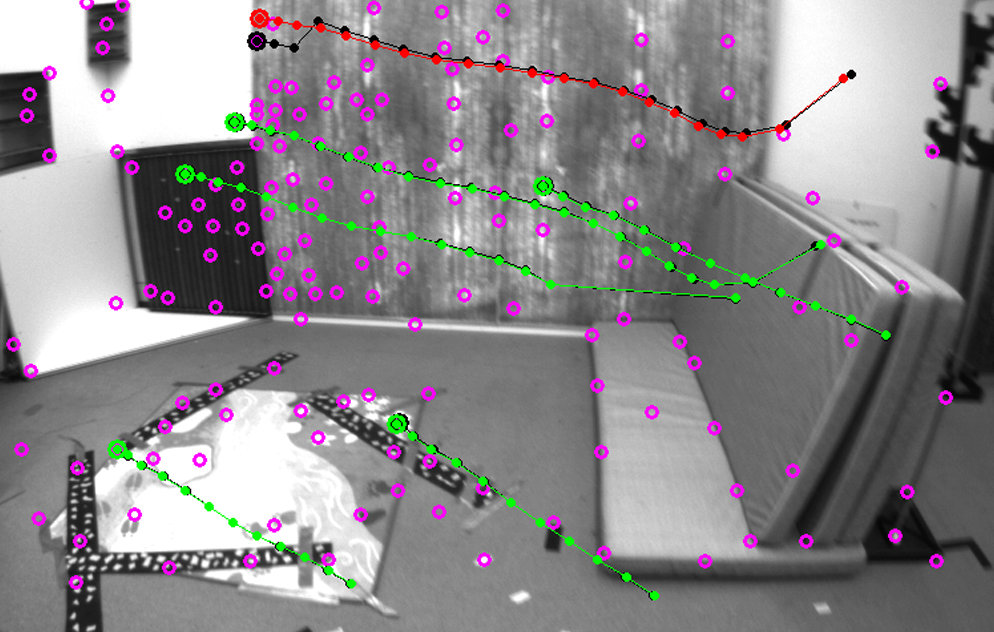}\\[3pt]
  \caption{Example feature tracking in a single EuRoC left frame. Selected feature tracks $\vy^j_S$ are shown in black (\cf\ \cref{sec:visual-update-track-selection}). The corresponding reprojections are drawn in green for successfully visual updates and red for tracks that failed the $\chi^2$ outlier test (\cf\ \cref{sec:visual-updates}). The end of the track with the larger circle matches the current frame, and the long gap between the two last key points in some tracks is a consequence of~\cref{eq:anti-track-reuse-criterion}. LK-tracked features that were not used on this frame are drawn in magenta.
  \label{fig:euroc-tracker-example}}
\end{figure}

The track selection logic, as well as some other aspects of our visual processing, are illustrated in \cref{fig:euroc-tracker-example}. The essence of this process is reducing computational load by focusing on the most informative visual features. The technique is inspired by the stochastic gradient descent method, and allows us to maintain good tracking performance with a very low number of $n_{\rm target} = 5$ active features per frame (see \cref{sec:experiments} for details).

\subsection{Camera model}
\noindent
Our method supports two different camera models: the radial-tangential distorted pinhole model and the Kannala--Brandt fisheye model~\cite{Kannala-Brandt-fisheye} with four radial parameters. We assume that the calibration parameters of the camera model are known and static.

The model $\phi_C: \R^2 \to \R^3$ for the camera $C \in \{ L, R \}$, maps the pixel coordinates of a feature to the ray bearing in $\R^3$, and its restriction to the unit sphere $S^2$ is invertible, $\phi_C^{-1} : S^2 \to \R^2$. We define the \emph{undistorted normalized pixel coordinates} for camera $C$ as
\begin{equation}
  \tilde{\vy}^{j,C}_l = \rho(\phi_C(\vy^{j,C}_l)),
\end{equation}
 where $\rho(x, y, z) = z^{-1} [x, y]^\T$ is the perspective projection.

When required, the extrinsic camera coordinates (as a camera-to-world transformation) are formed from the IMU coordinates $(\vp, \vq)$ in the VIO state as
\begin{equation}
  \label{eq:imu-to-camera}
  \MT^C(\vp, \vq) =
  {\footnotesize
    \left(
    \begin{matrix}
      \MR_C & \vp_C \\
       \\ & 1
    \end{matrix}
    \right)
  }
  =
  {\footnotesize
    \left(
    \begin{matrix}
      \MR^\T(\vq) & \vp \\
       \\ & 1
    \end{matrix}
    \right)
  }
  \MT_{C \mapsto {\rm IMU}},
\end{equation}
where $\MR(\vq)$ is the rotation matrix corresponding to the (world-to-IMU) quaternion $\vq$ and the IMU-to-camera transformation $\MT_{C \mapsto {\rm IMU}} \in \R^{4\times4}$ is determined for each camera $C$ in the calibration phase.

\subsection{Visual updates}
\label{sec:visual-updates}
\noindent
Following~\cite{PIVO}, our visual update is based on triangulation of feature tracks $\vy^j$ into 3D feature points $\vp^j$ and comparing their reprojections to the original features. In particular, this information is applied as an EKF update of the form
\begin{equation}
\label{eq:ekf-visual-update}
  \vh_{k,j}(\vx_{k|k-1,j-1}) = \vgamma_{k,j} \sim \mathcal N(\mbf{0}, \sigma^2_{\rm visu} \MI),
\end{equation}
with
\begin{equation}
  \label{eq:ekf-h}
  \vh_{k,j}(\vx) = \vr_S(\vp_S(\vx, \tilde{\vy}^j_{S}), \vx) - \tilde{\vy}^j_{S},
\end{equation}
where
\begin{equation}
  \label{eq:ekf-triangulation}
  \vp_S(\vx, \tilde{\vy}^j_{S}) = {\rm TRI}(\vpi^{(S)}, \tilde{\vy}^j_{S}) \in \R^3
\end{equation}
denotes triangulation using the selected subset $S = S(k,j)$ of the feature track points $\tilde{\vy}^j_{S}$ and the corresponding poses $\vpi^{(S)}$ in the VIO state.
The reprojections of the triangulated point on the corresponding frames are computed as
\begin{equation}
  \label{eq:ekf-reprojection}
  \vr_S(\vp^*, \vx) = [ \tilde\rho_C(\vp^*, \vpi^{(l)}) ]_{\substack{l \in S \\ C \in \mathcal C}},
\end{equation}
where $\mathcal C \subset \{L, R\}$ is the set of cameras (with two elements in stereo and one in mono) and
\begin{equation}
  \tilde\rho_C(\vp^*, \vp, \vq) = \rho\left(\MR_{C}^\T(\vq) \cdot (\vp^* - \vp_C)\right)
\end{equation}
projects a 3D point $\vp^*$ onto the normalized pixel coordinates of camera $C$ at pose $\vpi$ (\cf \cref{eq:imu-to-camera}).
We perform a $\chi^2$ outlier check before applying the EKF update corresponding to \cref{eq:ekf-visual-update}. In case of failure, we cancel the update and proceed to the next feature track as described in \cref{sec:visual-update-track-selection}.

Our method and PIVO differentiate from the other MSCKF variants with \cref{eq:ekf-h}: with tedious and repeated application of the chain rule of differentiation, one can compute the Jacobian $J_{\vh}$ with respect to the state $\vx$, which renders the linearization
\begin{equation}
  \vh_{k,j}(\vx) \approx J_{\vh,k,j}(\vx_0) (\vx - \vx_0) + \vh_{k,j}(\vx_0),
\end{equation}
directly usable in an EKF update step and the null space trick introduced in~\cite{MSCKF} becomes unnecessary.
We recompute the linearization for each update ($\vx_0 = \vx_{k|k-1,j-1}$).

\subsection{Triangulation}
\noindent
Our triangulation follows~\cite{PIVO}: the function ${\rm TRI}$ in \cref{eq:ekf-triangulation} uses a point $\vp^*_0$ computed from two camera rays as an initial value and then optimizes it by minimizing the reprojection error (\cf\ \cref{eq:ekf-reprojection})
\begin{equation}
\label{eq:reprojection-rmse}
  {\rm RMSE}_j(\vp^*, \vx) = \|\vr_S(\vp^*, \vx)\|
\end{equation}
using the Gauss--Newton algorithm. The entire optimization process needs to be differentiated with respect to the VIO state for computing the Jacobian for the EKF update (see \cref{supp:example-triangulation} for more details). In the stereo case, $\vp^*_0$ is computed using the most and least recent ray from the left camera only.
If the triangulation produces an invalid 3D point, \ie, behind any of the cameras, the feature track is rejected.

\subsection{Pose augmentation}
\label{sec:pose-augmentation}
\noindent
As in~\cite{MSCKF}, the pose trail $\vpi^{(\cdot)}$ in the VIO state is populated and updated in a process known as pose augmentation or \emph{stochastic cloning}~\cite{Roumeliotis-stochastic-cloning:2002}:
When a new camera frame is received at time $t_k$, a copy of the current, IMU-predicted, pose $\vpi^{(0)}$ is inserted into the slot $\vpi^{(1)}$, an older pose, $\vpi^{(d)}$ is discarded and the remaining poses are shifted accordingly. This operation can be performed as an EKF update step as described in~\cite{PIVO} or an EKF prediction step

  $\vx_{k+1|k} = \MA^{\rm aug}_d \vx_{k|k}$,
with
\begin{equation}
  \MA_d^{\rm aug} = \left(
  \begin{matrix}
    \MI_{n_1} & \\
    \MI_7 \\
    \mbf{0}_{\cdot \times n_1} & \MI_{7 (d - 1)} \\
    & & \mbf{0}_{\cdot\times7} & \MI_{7 (n_a - d)}
  \end{matrix}
  \right),
\end{equation}
where $n_1 := \dim(\vx) - 7\cdot n_a$.
The typical choice is always discarding the oldest pose, that is, $d = n_a$, which makes the pose trail effectively a FIFO queue.
However, by varying the discarded pose index, it is also possible to create more complex schemes that manage the pose trail as a generic $n_a$-slot memory.

\algrenewcommand{\algorithmiccomment}[1]{\hfill{\footnotesize \it \color{gray} #1}}
\begin{algorithm}[t]
  \small
  \caption{Hybrid VIO-SLAM}\label{algo:hybrid-vio-slam}
  \begin{algorithmic}[1]
    \Function{SlamTask}{$\MT_{\rm in}, (y_j)_{j \in U}, \mathcal I$}

      \State $\MT_{\rm slam} \gets $\Call{MatchGravityDir}{$\MT_{\rm slam} \MT_{\rm prev} \MT_{\rm in}^{-1}, \MT_{\rm in}$}\label{state:vio-glue}
      \State associate each $y_j$ with a map point $\MM_j$ \Comment{new or existing}
      \State initialize kf. candidate $\mathcal K = (\MT_{\rm slam}, K = (y_j)_{j \in U})$
      \If{\Call{KeyFrameDecision}{$\mathcal K$}}

      \State extend $K$ with more key points from the image $\mathcal I$
      \State compute ORB descriptors for all kps. $K$ \Comment{\cf~\cite{ORB}}
      \State match existing map points with $K$ \label{state:match-local-map}
      \State triangulate new map points \Comment{as in~\cite{ORB-SLAM}}
      \State deduplicate map points \Comment{as in~\cite{ORB-SLAM}}
      \State $\MT_{\rm slam} \gets $\Call{LocalBundleAdjustment}{$\mathcal K$} \label{state:local-ba}
      \State cull map points and key frames \Comment{as in~\cite{ORB-SLAM}}
      \EndIf
      \State $\MT_{\rm prev} \gets \MT_{\rm in}$ \Comment{stored for the next task, like $\MT_{\rm slam}$}
      \State \textbf{return} $\MT_{\rm slam}\MT_{\rm in}^{-1}$ \Comment{VIO $\to$ SLAM mapping}\label{state:return}
    \EndFunction
    \State $\MT_{\rm vio \to slam}, \MT_{\rm prev}, \MT_{\rm slam} \gets \MI_4$, $\mathcal F \gets $done \Comment{initialization} \label{state:control-begin}
    \For{\textbf{each} VIO frame $(\vpi_i^{(\cdot)}, (\vy^{j,L}_i)_{j \in U_i}, \mathcal I_i)$} \Comment{\cf~\cref{sec:vio-state}--\ref{sec:visual-update-track-selection}}
      \If{$i = 1\ ({\rm mod}\ N)$} \Comment{every $N$th frame}
        \State $\MT_{\rm vio \to slam} \gets $ \textbf{block} on $\mathcal F$ \label{state:wait-for-future}\Comment{wait for previous result}
        \State $\MT_{\rm in} \gets \MT^L(\vpi^{(N)}_i)$ \Comment{left camera pose in $N$th history slot}
        \State $\mathcal F \gets$ \textbf{start} \Call{SlamTask}{$\MT_{\rm in}, (\vy^{j,L}_i)_{j \in U_i}, \mathcal I_i$} \label{state:start-async} \Comment{async.}
      \EndIf
      \State \textbf{output} $\MT^{L,\rm out}_i \gets \MT_{\rm vio \to slam} \MT^L(\vpi^{(1)}_i)$ \Comment{latest pose $\vpi^{(1)}_i$}
    \EndFor \label{state:control-end}
  \end{algorithmic}
\end{algorithm}

\paragraph{Towers-of-Hanoi scheme}
We use
\begin{equation}
  \label{eq:fifo-and-towers-of-hanoi-scheme}
  d_i = \max(n_{\rm FIFO}, n_a - {\rm LSB}(i)),
\end{equation}
where ${\rm LSB}(i)$ denotes the least-significant zero bit index (0-based) of the integer $i$. This process combines a fixed-size $n_{\rm FIFO}$ part with a Towers-of-Hanoi backup scheme that increases the maximum age of the stored poses by adding (exponentially) increasing strides between them. It is also possible to vary $d_i$ dynamically based on, \eg, the number of tracked features in the corresponding camera frames. In particular, if a certain frame with corresponding historical pose $\vpi^{(l)}$ no longer shares any tracked feature points with the latest camera frame, we always discard it by setting $d = l$ instead of applying \cref{eq:fifo-and-towers-of-hanoi-scheme}.

The more complex scheme allows reducing the dimension of the state and computational load by using less poses $n_a$ more efficiently.

\subsection{Stationarity detection}\label{sec:stationarity-detection}
\noindent
The common important special case, where the tracked device is nearly stationary, requires some special attention in MSCKF-like methods. In particular, when the device is stationary, the pose augmentation schemes in \cref{sec:pose-augmentation} can quickly cause the pose trail to degenerate into $n_a$ (nearly) identical copies of a single point, which can destabilize the system. This concerns especially in the monocular scenario, as the triangulation baselines consequently approach zero.

We follow an approach also presented in~\cite{LARVIO}, where certain frames are classified as stationary, and not stored permanently in the pose trail. To this end, we evaluate the movement of the tracked features in pixel coordinates between consecutive frames. Namely, if
\begin{equation}
\label{eq:visual-stationarity-condition}
  m_k = \max_j \|\vy^{j,L}_k - \vy^{j,L}_{k-1} \| < m_{\min},
\end{equation}
for a certain fixed threshold $m_{\min}$, we perform a pose \emph{unaugmentation} operation as an EKF prediction step:
\begin{equation}
  \label{eq:pose-unaugmentation}
  \vx_{k+1|k} = (\MA_{n_a}^{\rm aug})^\T \vx_{k|k} +
  \left(\begin{matrix}
    \mbf{0}_{\dim(\vx) - 7} & \\
    & \vepsilon_u
  \end{matrix}\right),
\end{equation}
where $\vepsilon_u \sim \mathcal N(\mbf{0}, \sigma_u^2\MI_7)$ with a large variance (\eg, $\sigma_u \approx 10^6$). This causes the previously augmented pose to be discarded (after it has been used for a visual update) and, as a result, most of the frames remain in the pose trail as long as the device remains stationary and \cref{eq:visual-stationarity-condition} holds.

\subsection{SLAM module}
\label{sec:slam-module}
\noindent
On a high level, our method consists of two loosely coupled modules: the filtering-based VIO module, which is described in previous sections, and an optional, optimization-based SLAM module, which uses VIO as an input. We used OpenVSLAM~\cite{OpenVSLAM}, a re-implementation of the ORB-SLAM2~\cite{ORB-SLAM2} method, as the basis for the implementation. Consequently, many of the details of our SLAM module coincide with ORB-SLAM2 or its predecessor, ORB-SLAM~\cite{ORB-SLAM}. We describe these parts of the system briefly and refer the reader to the aforementioned works for details.

\paragraph{SLAM map structure}
A sparse SLAM map consists of \emph{key frames} and \emph{map points}, which are observed as 2D \emph{key points} in one or more key frames.
Equally to ORB-SLAM, our map point structure includes the viewing direction, valid distance range, and an ORB descriptor, while the \emph{key frame} consists of a list of key points and a camera pose.

\paragraph{ORB detection and matching}
\label{sec:orb-matching}
Unlike ORB-SLAM2, we only consider the data in the left camera frames in the stereo case for simplicity, even though our VIO module uses data from both cameras.
Each key point is associated with an ORB descriptor, which, in ORB-SLAM, are computed using a multi-pyramid-level FAST detector (\cf\ \cite{ORB}). In addition to this, we use the pixel coordinates of the Lucas--Kanade tracker features (\cf\ \cref{sec:feature-tracking}) as key points and compute their ORB descriptors on a single pyramid level.

New matches between key points are also searched from the $n_{\rm matching}$ key frames spatially closest to the current key frame.
As in ORB-SLAM, this is conducted both with \emph{3D matching}, where an existing map point is reprojected to the target key frame, and with 2D ORB matching, where the descriptors in two key frames are compared. The latter approach can be used to create new, previously untriangulated map points.
Previously visited areas can be recognized here without a separate \emph{loop closure} procedure (\cf\ \cite{ORB-SLAM3}, §VII) when the accumulated error is low enough.
In the SLAM module, we use linear triangulation for new map points.

\paragraph{VIO integration}\label{sec:vio-integration}
A high-level structure of our hybrid VIO--SLAM approach is given in \cref{algo:hybrid-vio-slam}, where lines \ref{state:control-begin}--\ref{state:control-end} describe a simple and efficient parallel scheme: the VIO state is sent to the SLAM module, which outputs a VIO-to-SLAM coordinate mapping. The result is read asynchronously on the next key frame candidate, which we add every $N = 8$ frames. The returned coordinate mapping is not required to match the latest pose and we input a fixed-delay-smoothed VIO pose $\vpi^{(N)}$ to SLAM, while outputting an undelayed pose on each input frame, using the most recent available $\MT_{\rm vio \to slam}$.

We initialize the new key frame at a pose transformed using recent key frame and input poses as shown on \ref{state:vio-glue}, where \textsc{MatchGravityDir}$(\MT, \MT_{\rm in})$ ensures that $\MT = {\rm rotate}_z(\theta)\MT_{\rm in}$ for some $\theta$, that is, the gravity direction in the initial key frame pose matches that of the VIO input.
The \textsc{KeyFrameDecision} passes if the distance from the previous key frame exceeds a fixed threshold (15cm), or if less 70\% of the feature tracks are covisible in it.

\paragraph{Bundle adjustment} Local bundle adjustment (\cf\ \cite{ORB-SLAM}) is performed on $n_{\rm BA}$ nearest neighbor key frames (by Euclidean distance) of current key frame.
In addition, we use the relative input pose changes $\MT_{\rm in, i}^{-1}\MT_{\rm in, i - N}$ from VIO as extra penalty terms between consecutive key frames to limit the deviation between the SLAM and VIO trajectory shapes. Our penalty weights for both position and orientation are inversely proportional to the time interval $t_i - t_{i - N}$.

\paragraph{Post-processing}
As the \emph{post-processed} trajectory in \cref{sec:experiments}, we use the final positions of the key frames and interpolate between them using the online VIO trajectory to produce a pose estimate for each input frame.

\begin{table}[h!]
  \caption{System parameters.}
  \label{tbl:parameters}
  \vspace{4pt}
  \footnotesize
  \setlength{\tabcolsep}{3.5pt}
  \begin{tabular}{P{3em}l|cccc}
  & Parameter & \shortstack{Fast\\VIO} & \shortstack{Normal\\VIO} & \shortstack{Normal\\SLAM} & \shortstack{Post-proc.\\SLAM} \\
  \toprule
  \multirow{2}{*}{\shortstack{feature\\detector}} &type& FAST & GFTT  & GFTT & GFTT \\
  &subpixel adjustment & no & yes & yes & yes \\
  \cline{1-2}
  \multirow{4}{*}{\shortstack{feature\\tracker}}
    & max.\ features (stereo) & 70 & 200 & 200 & 200 \\
    & max.\ features (mono) & 100 & 200 & 200 & 200 \\
    & max.\ itr. & 8 & 20 & 20 & 20\\
    & window size & 13 & 31 & 31 & 31 \\
  \cline{1-2}
  \multirow{3}{*}{\shortstack{visual\\updates}}
  & $n_a$ (\cref{sec:vio-state}) & 6 & 20 & 20 & 20 \\
  & $n_{\rm target}$ (\cref{sec:visual-update-track-selection})  & 5 & 20 & 20 & 20\\
  & $n_{\rm FIFO}$ (\cref{eq:fifo-and-towers-of-hanoi-scheme}) & 2 & 17 & 17 & 17 \\
\cline{1-2}
  \multirow{2}{*}{SLAM} & $n_{\rm BA}$ & -- & -- & 20 & 100 \\
  & $n_{\rm matching}$ & -- & -- & 20 & 50 \\
  \bottomrule
  \end{tabular}
\end{table}

\begin{figure}[t!]
     \centering
     \begin{tikzpicture}
      \node[anchor=south west,inner sep=0] (image) at (0,0) {\includegraphics[width=\linewidth]{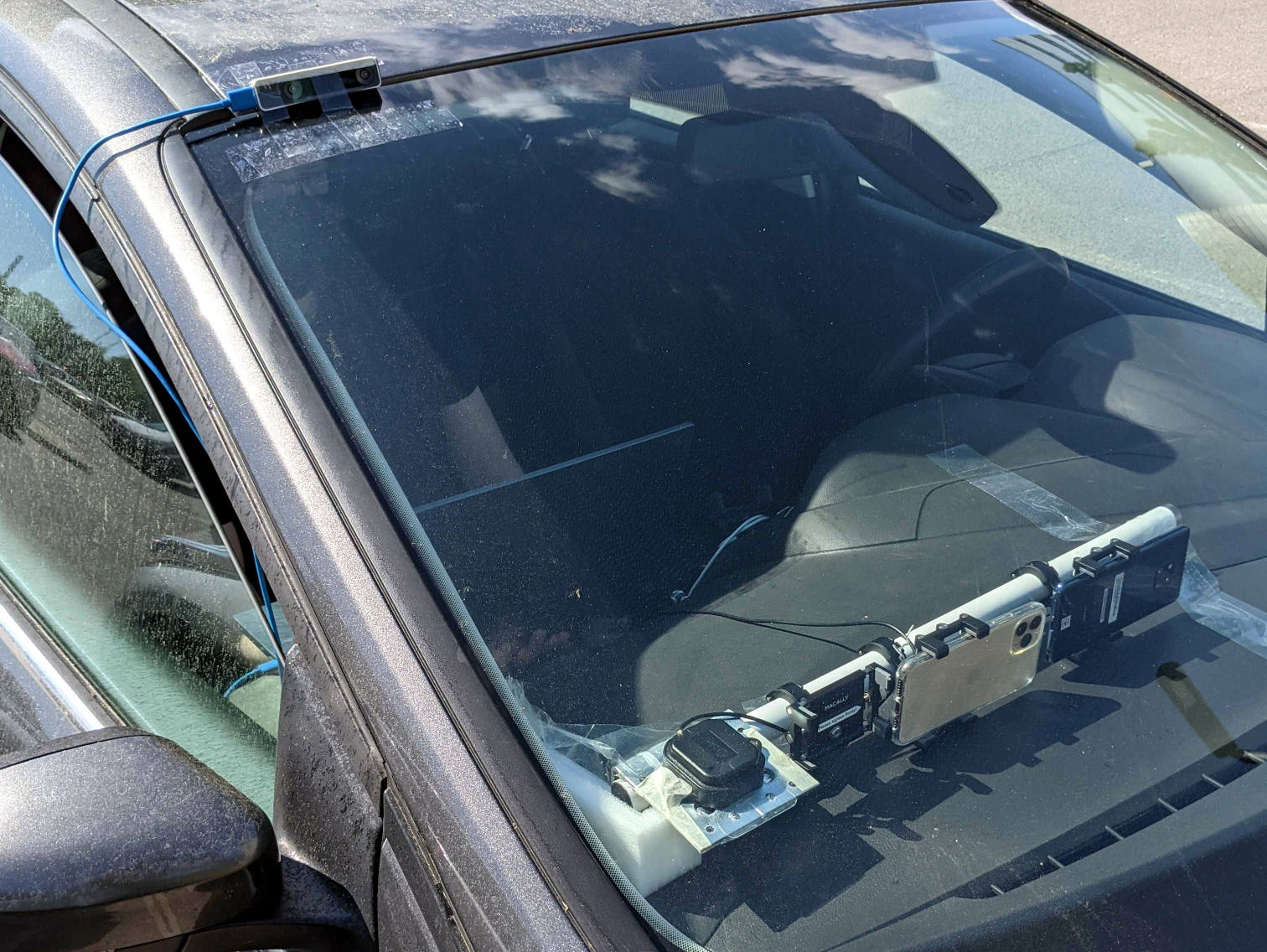}};
      \begin{scope}[x={(image.south east)},y={(image.north west)}]
          \fill [gray, rounded corners=2pt, opacity=.8] (0.61,0.68) rectangle (1,1);
           \def\annotationcolor{yellow}
           \node [anchor=west, \annotationcolor] (legendRS) at (0.63,0.95) {\small RealSense T265};
           \node [anchor=west, \annotationcolor] (legendRTK) at (0.63,0.88) {\small u-blox RTK-GNSS};
           \node [anchor=west, \annotationcolor] (legendIPhone) at (0.63,0.81) {\small iPhone 11 Pro};
           \node [anchor=west, \annotationcolor] (legendHuawei) at (0.63,0.74) {\small Huawei Mate 20 Pro};
          \draw [-latex, thick, \annotationcolor] (legendRS) to[out=180, in=0] (0.33,0.91);
          \draw [-latex, thick, \annotationcolor] (legendRTK) to[out=180, in=90] (0.55,0.25);
          \draw [-latex, thick, \annotationcolor] (legendIPhone) to[out=180, in=120] (0.75,0.35);
          \draw [-latex, thick, \annotationcolor] (legendHuawei) to[out=270, in=120] (0.87,0.45);
      \end{scope}
      \end{tikzpicture}\vspace{3pt}
     \caption{Car setup: GNSS is used as ground truth. Other devices record their proprietary VISLAM output (RealSense, ARKit on iOS 14.3, or ARCore 1.21) and its inputs (IMU \& cameras).\label{fig:car-measurement-setup}}
\end{figure}
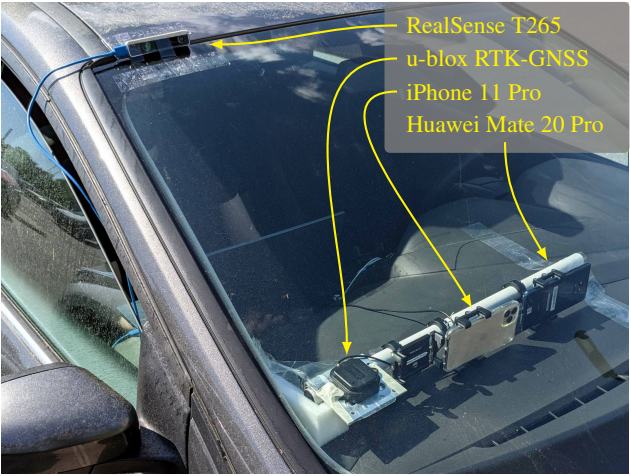

\def\ourtablefontsize{\scriptsize}

\begin{table*}[!t]
  \caption{EuRoC MAV benchmark (RMS ATE metric with $SE(3)$ alignment, in meters).}
  \vspace{2pt}
  \label{tbl:euroc-combined}
  \centering
  \ourtablefontsize
  \renewcommand{\tabcolsep}{8.2pt}
  \begin{tabular}{>{\raggedright}p{0.75em}|p{0.5em}|p{7em}|P{2.2em}P{2.2em}P{2.2em}P{2.2em}P{2.2em}P{2.2em}P{2.2em}P{2.2em}P{2.2em}P{2.2em}P{2.2em}|P{2.2em}}
 &  & Method & MH01 & MH02 & MH03 & MH04 & MH05 & V101 & V102 & V103 & V201 & V202 & V203 & Mean\\
\toprule
\multirow{10}{*}{\rotatebox{90}{online}} & \multirow{4}{*}{\rotatebox{90}{stereo}} & OKVIS & 0.23 & 0.15 & 0.23 & 0.32 & 0.36 & \textbf{0.04} & 0.08 & 0.13 & 0.10 & 0.17 & \textit{--} & 0.18\\
 &  & VINS-Fusion & 0.24 & 0.18 & 0.23 & 0.39 & 0.19 & 0.10 & 0.10 & 0.11 & 0.12 & 0.10 & \textit{--} & 0.18\\
 &  & BASALT & \textbf{0.07} & \textbf{0.06} & \textit{0.07} & \textit{0.13} & \textit{0.11} & \textbf{0.04} & \textit{0.05} & \textit{0.10} & \textbf{0.04} & \textit{0.05} & \textit{--} & \textit{0.072}\\
 &  & Ours\textsuperscript{(1)} & \textit{0.088} & \textit{0.08} & \textbf{0.038} & \textbf{0.071} & \textbf{0.11} & 0.044 & \textbf{0.035} & \textbf{0.04} & \textit{0.075} & \textbf{0.041} & \textbf{0.052} & \textbf{0.061}\\
\cline{2-15}
 & \multirow{5}{*}{\rotatebox{90}{mono}} & OKVIS & 0.34 & 0.36 & 0.30 & 0.48 & 0.47 & 0.12 & 0.16 & 0.24 & 0.12 & 0.22 & -- & 0.28\\
 &  & PIVO & -- & -- & -- & -- & -- & 0.82 & -- & 0.72 & 0.11 & 0.24 & \textit{0.51} & 0.48\\
 &  & VINS-Fusion & \textit{0.18} & 0.09 & 0.17 & \textit{0.21} & \textit{0.25} & \textbf{0.06} & 0.09 & 0.18 & 0.06 & 0.11 & -- & 0.14\\
 &  & VI-DSO & \textbf{0.06} & \textbf{0.04} & \textbf{0.12} & \textbf{0.13} & \textbf{0.12} & \textbf{0.06} & \textit{0.07} & \textit{0.10} & \textbf{0.04} & \textbf{0.06} & -- & \textbf{0.08}\\
 &  & Ours\textsuperscript{(2)} & 0.19 & \textit{0.066} & \textit{0.12} & 0.21 & 0.31 & 0.069 & \textbf{0.061} & \textbf{0.08} & \textit{0.052} & \textit{0.089} & \textbf{0.13} & \textit{0.13}\\
\cline{1-15}
\multirow{10}{*}{\rotatebox{90}{post-processed}} & \multirow{6}{*}{\rotatebox{90}{stereo}} & OKVIS & 0.160 & 0.220 & 0.240 & 0.340 & 0.470 & 0.090 & 0.200 & 0.240 & 0.130 & 0.160 & 0.290 & 0.23\\
 &  & VINS-Fusion & 0.166 & 0.152 & 0.125 & 0.280 & 0.284 & 0.076 & 0.069 & 0.114 & 0.066 & 0.091 & 0.096 & 0.14\\
 &  & BASALT & 0.080 & 0.060 & 0.050 & 0.100 & \textit{0.080} & 0.040 & \textit{0.020} & \textit{0.030} & \textbf{0.030} & \textit{0.020} & -- & 0.051\\
 &  & Kimera & 0.080 & 0.090 & 0.110 & 0.150 & 0.240 & 0.050 & 0.110 & 0.120 & 0.070 & 0.100 & 0.190 & 0.12\\
 &  & ORB-SLAM3 & \textbf{0.037} & \textit{0.031} & \textbf{0.026} & \textit{0.059} & 0.086 & \textbf{0.037} & \textbf{0.014} & \textbf{0.023} & 0.037 & \textbf{0.014} & \textbf{0.029} & \textbf{0.036}\\
 &  & Ours\textsuperscript{(3)} & \textit{0.048} & \textbf{0.028} & \textit{0.037} & \textbf{0.056} & \textbf{0.066} & \textit{0.038} & 0.035 & 0.037 & \textit{0.031} & 0.029 & \textit{0.044} & \textit{0.041}\\
\cline{2-15}
 & \multirow{4}{*}{\rotatebox{90}{mono}} & VINS-Mono & 0.084 & 0.105 & 0.074 & \textit{0.122} & 0.147 & 0.047 & 0.066 & 0.180 & 0.056 & 0.090 & 0.244 & 0.11\\
 &  & VI-DSO & 0.062 & \textbf{0.044} & 0.117 & 0.132 & \textit{0.121} & 0.059 & 0.067 & 0.096 & \textbf{0.040} & 0.062 & 0.174 & 0.089\\
 &  & ORB-SLAM3 & \textbf{0.032} & 0.053 & \textbf{0.033} & \textbf{0.099} & \textbf{0.071} & \textit{0.043} & \textbf{0.016} & \textbf{0.025} & \textit{0.041} & \textbf{0.015} & \textbf{0.037} & \textbf{0.042}\\
 &  & Ours\textsuperscript{(4)} & \textit{0.056} & \textit{0.048} & \textit{0.066} & 0.13 & 0.13 & \textbf{0.039} & \textit{0.044} & \textit{0.065} & 0.044 & \textit{0.047} & \textit{0.056} & \textit{0.066}\\
\cline{1-15}
\end{tabular}

\end{table*}

\begin{table*}[!ht]
  \caption{Different configurations (\cf\ \cref{tbl:parameters}) of \ourmethodname. The symbol $\smallsetminus$ marks features removed from a baseline configuration (topmost in the same box). The row labeled \emph{PIVO baseline} represents our reimplentation of~\cite{PIVO}, obtained by disabling \emph{all} novel features mentioned in this table from \emph{Normal VIO}. Average frame processing times are given for a high-end desktop (\emph{Ryzen}) and an embedded (\emph{R-Pi}) CPU.}
  \vspace{2pt}
  \label{tbl:euroc-our-variants}
  \centering
  \ourtablefontsize

  \renewcommand{\tabcolsep}{6.5pt}
  \begin{tabular}{>{\raggedright}P{1em}|P{1em}|p{8em}P{2em}P{2em}P{2em}P{2em}P{2em}P{2em}P{2em}P{2em}P{2em}P{2em}P{2em}|P{2em}|P{2em}P{2em}}
 &  &  &  &  &  &  &  &  &  &  &  &  &  &  & \multicolumn{2}{m{4.8em}}{frame (ms)}\\
 &  & Method & MH01 & MH02 & MH03 & MH04 & MH05 & V101 & V102 & V103 & V201 & V202 & V203 & Mean & Ryzen & R-Pi\\
\toprule
\multirow{15}{*}{\rotatebox{90}{online}} & \multirow{5}{*}{\rotatebox{90}{stereo}} & Normal SLAM\textsuperscript{(1)} & 0.09 & 0.08 & 0.04 & 0.07 & 0.11 & 0.04 & 0.03 & 0.04 & 0.08 & 0.04 & 0.05 & 0.061 & 32 & -\\
\cline{3-17}
 &  & Normal VIO & 0.08 & 0.07 & 0.15 & 0.10 & 0.10 & 0.06 & 0.06 & 0.09 & 0.05 & 0.04 & 0.12 & 0.084 & 32 & -\\
 &  & $\smallsetminus$ \cref{eq:anti-track-reuse-criterion} & 0.08 & 0.09 & 0.12 & 0.14 & 0.14 & 0.05 & 0.06 & 0.13 & 0.05 & 0.06 & 0.14 & 0.095 & 99 & -\\
\cline{3-17}
 &  & Fast VIO & 0.26 & 0.09 & 0.15 & 0.10 & 0.18 & 0.11 & 0.05 & 0.10 & 0.07 & 0.07 & 0.12 & 0.12 & 8.3 & 49\\
 &  & $\smallsetminus$ \cref{eq:fifo-and-towers-of-hanoi-scheme} & 0.30 & 0.28 & 0.22 & 0.17 & 0.18 & 0.09 & 0.05 & 0.14 & 0.10 & 0.11 & 0.15 & 0.16 & 9 & 53\\
\cline{2-17}
 & \multirow{10}{*}{\rotatebox{90}{mono}} & Normal SLAM\textsuperscript{(2)} & 0.19 & 0.07 & 0.12 & 0.21 & 0.31 & 0.07 & 0.06 & 0.08 & 0.05 & 0.09 & 0.13 & 0.13 & 16 & -\\
\cline{3-17}
 &  & Normal VIO & 0.24 & 0.14 & 0.33 & 0.26 & 0.39 & 0.06 & 0.07 & 0.11 & 0.05 & 0.15 & 0.13 & 0.18 & 16 & -\\
 &  & $\smallsetminus$ RANSAC & 0.42 & 0.17 & 0.29 & 0.27 & 0.42 & 0.08 & 0.08 & 0.15 & 0.05 & 0.14 & 0.13 & 0.2 & 15 & -\\
 &  & $\smallsetminus$ \cref{eq:imu-bias-propagation} & 0.31 & 0.23 & 0.22 & 0.42 & 0.46 & 0.11 & 0.21 & 0.31 & 0.15 & 0.23 & 9.02 & 1.1 & 16 & -\\
 &  & $\smallsetminus$ \cref{eq:longer-than-median-tracks} & 0.25 & 0.43 & 0.27 & 0.22 & 0.40 & 0.06 & 0.08 & 0.14 & 0.06 & 0.12 & 0.14 & 0.2 & 15 & -\\
 &  & $\smallsetminus$ \cref{sec:stationarity-detection} & 4.95 & 2.70 & 0.34 & 0.34 & 0.45 & 0.25 & 0.23 & 0.51 & 0.49 & 0.75 & 0.17 & 1 & 16 & -\\
 &  & $\smallsetminus$ \cref{eq:anti-track-reuse-criterion} & 0.22 & 0.17 & 0.24 & 0.25 & 0.38 & 0.06 & 0.07 & 0.15 & 0.06 & 0.09 & 0.17 & 0.17 & 36 & -\\
 &  & PIVO baseline & 0.38 & 0.24 & 0.23 & 0.40 & 0.43 & 0.22 & 0.28 & 0.39 & 0.32 & 0.39 & 3.25 & 0.59 & 30 & -\\
\cline{3-17}
 &  & Fast VIO & 0.37 & 0.35 & 0.43 & 0.30 & 0.36 & 0.10 & 0.09 & 0.12 & 0.08 & 0.18 & 0.12 & 0.23 & 5.3 & 33\\
 &  & $\smallsetminus$ \cref{eq:fifo-and-towers-of-hanoi-scheme} & 0.95 & 0.73 & 0.71 & 0.48 & 0.67 & 0.20 & 0.11 & 0.13 & 0.11 & 0.18 & 0.17 & 0.4 & 5.4 & 33\\
\cline{1-17}
\multicolumn{2}{P{2.2em}|}{\multirow{2}{*}{post-pr.}} & Stereo SLAM\textsuperscript{(3)} & 0.05 & 0.03 & 0.04 & 0.06 & 0.07 & 0.04 & 0.03 & 0.04 & 0.03 & 0.03 & 0.04 & 0.041 & 52 & -\\
\multicolumn{2}{P{2.2em}|}{} & Mono SLAM\textsuperscript{(4)} & 0.06 & 0.05 & 0.07 & 0.13 & 0.13 & 0.04 & 0.04 & 0.06 & 0.04 & 0.05 & 0.06 & 0.066 & 43 & -\\
\cline{1-17}\end{tabular}

\end{table*}

\section{Experiments}
\noindent\label{sec:experiments}%
We compared our approach to the current-state-of-the-art~\cite{ORB-SLAM3,VINS-Mono,Kimera,BASALT,VI-DSO} in three academic benchmarks. Two baseline methods, OKVIS~\cite{OKVIS}, for which results are reported in all of these, and PIVO~\cite{PIVO}, the most similar method, were also included. More comprehensive comparisons including older methods and visual-only approaches can be found in~\cite{ORB-SLAM3} and \cite{BASALT}, which are also our primary sources of the results for other methods in \cref{tbl:euroc-combined,tbl:tum-vi-room}.

\subsection{EuRoC MAV}
\label{sec:experiments-euroc}
\noindent
\cref{tbl:euroc-combined} gives our results for the EuRoC MAV~\cite{EuRoC}. Similarly to~\cite{BASALT}, we clearly separate the \emph{online} and \emph{post-processed} cases. The former corresponds to real-time estimation of the current device pose using the data seen so far. The latter, also called \emph{mapping} mode, aims to produce an accurate post-processed trajectory using all data in the sequence. In Bayesian terms, they are the \emph{filtered} and \emph{smoothed} solutions, respectively.

Our approach yields state-of-the-art performance in all categories: monocular and stereo, as well as online and post-processed. Furthermore, we outperform BASALT~\cite{BASALT} in the online stereo category, and consequently, report the best real-time accuracy ever published for EuRoC. The authors of ORB-SLAM3~\cite{ORB-SLAM3}, the best method in the post-processing categories, do not report online results, but according to our experiments with the published source code (\cref{sec:orb-slam-comparison}), the online performance is not good (\cf\ \cref{fig:teaser-euroc-uncertainty}).

\paragraph{Parameter variations and timing} In \cref{tbl:euroc-our-variants}, we examine the accuracy and computational load with four different configurations detailed in \cref{tbl:parameters}. In addition, we measure the effect of the improvements presented in \cref{sec:methods}. Removing RANSAC, IMU bias random walks \cref{eq:imu-bias-propagation}, track selection logic (\cref{eq:longer-than-median-tracks}), stationarity detection \cref{sec:stationarity-detection} or \cref{eq:fifo-and-towers-of-hanoi-scheme} results in measurable reductions in accuracy.
We also removed all novel features simultaneously and this configuration represents our reimplementation of the PIVO method.
The average performace of the reimplementation is comparable to the numbers published in~\cite{PIVO} and reproduced in \cref{tbl:euroc-combined}, but the individual numbers are not identical. We presume this is mostly due to random variation and minor differences in unpublished implementation details.

The computational load is evaluated on two different machines: A high-end desktop computer with an AMD Ryzen 9 3900X processor, and a Raspberry Pi 4 with an ARM Cortex-A72 processor for simulating an embedded system. Both systems run Ubuntu Linux 20.04 and the maximum RAM consumption in the EuRoC benchmark was below 500 MB.
For the VIO-only (non-SLAM) variants, we measure single-core performance so that our method runs in a single thread, but auxiliary threads are used for decoding the EuRoC image data from disk, simulating a real-time use case where this data is processed online. In the SLAM case, we use two processing threads: one for SLAM and one for VIO, as described in \cref{algo:hybrid-vio-slam}. The EuRoC camera data is recorded at 20 FPS and thus values less than 50~ms per frame correspond to real-time performance, which is achieved in all unablated (\ie\ including \cref{eq:anti-track-reuse-criterion}) online cases on the desktop CPU and the \emph{fast VIO} configurations on the embedded processor.

\begin{table}
  \centering
  \caption{TUM VI (Room), post-processed, RMSE in meters.}
  \vspace{3pt}
  \label{tbl:tum-vi-room}
  \ourtablefontsize
  \setlength{\tabcolsep}{5pt}
  \renewcommand{\arraystretch}{.75}
  \begin{tabular}{>{\raggedright}p{6em}|P{1.8em}P{1.8em}P{1.8em}P{1.8em}P{1.8em}P{1.8em}|P{1.8em}}
Method & R1 & R2 & R3 & R4 & R5 & R6 & Mean\\
\toprule
OKVIS & 0.06 & 0.11 & 0.07 & 0.03 & 0.07 & 0.04 & 0.063\\
BASALT & 0.09 & 0.07 & 0.13 & 0.05 & 0.13 & 0.02 & 0.082\\
ORB-SLAM3 & \textbf{0.008} & \textbf{0.012} & \textit{0.011} & \textbf{0.008} & \textbf{0.010} & \textbf{0.006} & \textbf{0.009}\\
Ours & \textit{0.016} & \textit{0.013} & \textbf{0.011} & \textit{0.013} & \textit{0.02} & \textit{0.01} & \textit{0.014}\\
\cline{1-8}\end{tabular}

  \vspace*{-1em}
\end{table}

Even though the SLAM module increases accuracy in both monocular and stereo cases, the VIO-only mode also has very good performance compared to other approaches. In particular, by comparing the results to \cref{tbl:euroc-combined}, we note that our VIO-only stereo method outperforms VINS-Fusion even with the \emph{fast} settings. An example trajectory with this configuration is shown in \cref{fig:teaser-euroc-uncertainty}, which also illustrates how the EKF covariance can be used for uncertainty quantification with essentially no extra computational cost.

\label{sec:experiments-timing}

\begin{figure*}[!t]
     \centering
     \hfill
     \begin{subfigure}[b]{0.37\linewidth}
         \centering
         \includegraphics[width=\textwidth]{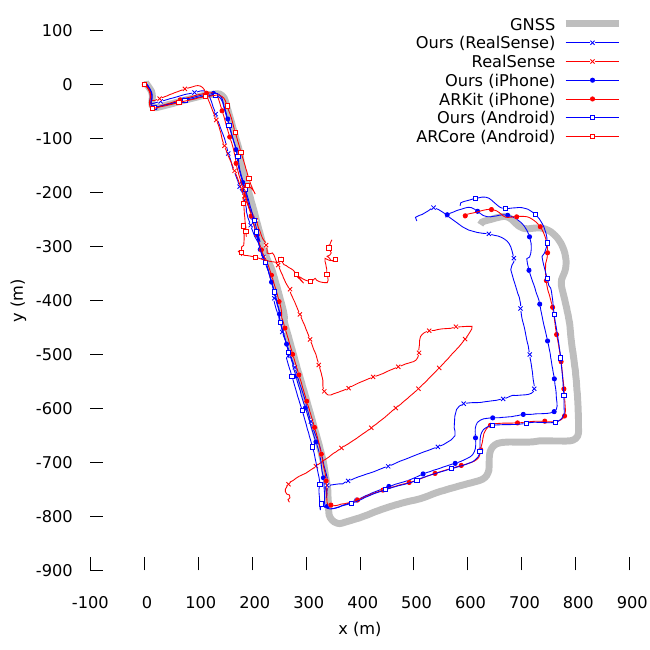}
         \caption{Vehicular (\cf\ \cref{fig:car-measurement-setup}). Maximum velocity ${\sim}$40km/h.\label{subfig:car-result}}
     \end{subfigure}
     \hfill
     \begin{subfigure}[b]{0.37\linewidth}
         \centering
         \includegraphics[width=\textwidth]{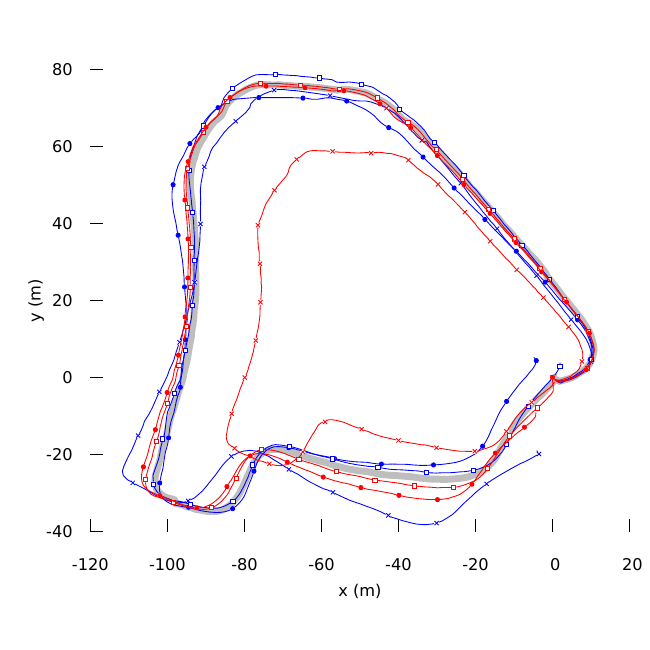}
         \caption{Walking\label{subfig:walking-result}}
     \end{subfigure}
     \hfill
     \vspace{0.5em}
     \caption{Comparison to commercial solutions.
     The lines with the same symbol use the same device and input data: RealSense T265 ($\times$), iPhone 11 Pro ($\bullet$), or Huawei Mate 20 Pro ({\tiny $\Box$}). Blue line is our result and red is a commercial solution on the same device.\label{fig:commercial-comparison}}
     \vspace*{-.5em}
\end{figure*}

\begin{table}
  \centering
  \caption{SenseTime Benchmark, online, RMSE in millimeters.}
  \vspace{3pt}
  \label{tbl:sensetime}
  \ourtablefontsize
  \setlength{\tabcolsep}{5.3pt}
  \renewcommand{\arraystretch}{.75}
  \begin{tabular}{>{\raggedright}p{6.7em}|P{1.1em}P{1.1em}P{1.1em}P{1.1em}P{1.1em}P{1.1em}P{1.1em}P{1.1em}|P{1.1em}}
Method & A0 & A1 & A2 & A3 & A4 & A5 & A6 & A7 & Mean\\
\toprule
OKVIS & 71.7 & 87.7 & 68.4 & 22.9 & 147 & 77.9 & 63.9 & 47.5 & 73.4\\
VINS-Mono & 63.4 & 80.7 & 74.8 & \textit{20} & \textit{18.7} & 42.5 & \textit{26.2} & 18.2 & 43.1\\
SenseSLAM {\tiny v1.0} & \textit{59} & \textit{55.1} & \textit{36.4} & \textbf{17.8} & \textbf{15.6} & \textbf{34.8} & \textbf{20.5} & \textbf{10.8} & \textit{31.2}\\
Ours & \textbf{49.9} & \textbf{30} & \textbf{36} & 22.2 & 19.6 & \textit{37.8} & 29.3 & \textit{17.3} & \textbf{30.3}\\
\cline{1-10}\end{tabular}

  \vspace*{-1em}
\end{table}

\subsection{TUM VI and SenseTime VISLAM}
\label{sec:experiments-tum-vi}
\noindent
We also evaluate our method on the \emph{room} subset of the TUM VI benchmark~\cite{TUMdataset} (\cref{tbl:tum-vi-room}) and the SenseTime VISLAM Benchmark~\cite{SenseTime} (\cref{tbl:sensetime}) which measures the performance of monocular VISLAM with smartphone data.
Both benchmarks measure post-processed SLAM performance using the RMS-ATE-$SE(3)$ metric.
In TUM VI Room, we rank second, after ORB-SLAM3. In the SenseTime benchmark, we outperform the authors' own proprietary method, \emph{Sense\-SLAM}, on the average. More parameter configurations are presented in \cref{sec:additional-experiments-ablation}.

\subsection{Commercial comparison dataset}
\noindent\label{sec:commercial-comparison}%
To evaluate the performance of our method compared to (consumer-grade) commercial solutions, we collected a custom dataset using the equipment depicted in \cref{fig:car-measurement-setup}. Each of the devices
features a commercial VISLAM algorithm, whose outputs can be recorded, together with the camera and IMU data the device observes. This allows us to compare the accuracy of our approach to the outputs of each commercial method with the same input.

\cref{fig:commercial-comparison} shows the output trajectories of the experiment for two different sequences: \cref{subfig:car-result} shows a vehicular test, where devices were attached to a car, exactly as shown in \cref{fig:car-measurement-setup}. In \cref{subfig:walking-result}, the same devices were rigidly attached to a short rod and carried by a walking person.

While all methods performed relatively well in the walking sequence, this is not the case in the more challenging vehicular test, which is not officially supported by any of the tested devices. However, our method (and notably, also ARKit) are able to produce stable tracking in all cases. We also clearly outperform Intel RealSense
in both sequences.

\section{Discussion and conclusions}
\label{sec:discussion}
\noindent%
We demonstrated how the PIVO framework could be extended to stereoscopic data and improved into a high-performance independent VIO method. Furthermore, we demonstrated a novel scheme for extending it with a parallel, loosely-coupled SLAM module. The resulting hybrid method outperforms the previous state-of-the-art in real-time stereo tracking.

The measurement of VIO-only performance is also relevant since the relative value of different VISLAM capabilities are dependent on the use case. For example, in vehicular setting where GNSS-VIO fusion is utilized to perform tracking during GPS breaks, \eg, in tunnels; loop closures or local visual consistency may be irrelevant compared to uncertainty quantification and long-range accuracy. In this case, we presume that a light-weight VIO solution is more suitable than full VISLAM. We also demonstrated the feasibility of our method for vehicular tracking.

With slight trade-off for accuracy, real-time performance was demonstrated on a Raspberry Pi without the use of GPU, VPU or ISP resources, which could further improve the speed and energy consumption of visual processing. The alternative approaches report similar real-time accuracy only on high-end desktop CPUs.

Note that several aspects of our VIO are simplified compared to other recent publications. In particular, the initialization presented in \cref{sec:imu-propagation} is extremely simple compared to the intricate mechanisms in~\cite{ORB-SLAM3} and~\cite{ORB-SLAM3-initialization}; we do not use the First--Estimate Jacobian methodology~\cite{first-estimate-jacobians-paper}, nor model orientations as probability distributions on the $SO(3)$ manifold~\cite{Forster+Carlone+Dellaert+Scaramuzza:2017}. Implementing some of these techniques could further improve the accuracy of this approach.

Similarly to BASALT, our SLAM module lacks a separate loop closure procedure, since on the tested datasets, the low online drift could always be corrected in other SLAM steps. However, a loop closure approach similar to \cite{ORB-SLAM} could be valuable in  challenging, large-scale benchmarks.

For an open-source implementation of the HybVIO method, see
 \url{https://github.com/SpectacularAI/HybVIO}.

\ifwacvfinal
\paragraph{Acknowledgments} We would like to thank Johan Jern for his contribution to the early versions of our SLAM module, and Iurii Mokrii for his contribution to our stereo code base.
\fi

{\small
\bibliographystyle{ieee_fullname}
\bibliography{bibliography}
}

\clearpage

\appendix
\twocolumn[\vspace*{5em}\centering\Large\bf%
Appendix
\vspace*{4em}]

\setcounter{table}{0}
\renewcommand{\thetable}{A\arabic{table}}%
\setcounter{figure}{0}
\renewcommand{\thefigure}{A\arabic{figure}}%
\setcounter{equation}{0}
\renewcommand{\theequation}{A\arabic{equation}}%

\section{Examples and experiment details}
\label{supp:example-triangulation}\noindent%
\paragraph{Differentiation example} Consider the case where the triangulation is performed using two poses $\vpi^{(1)}$, $\vpi^{(2)}$ in the stereo setup:
\begin{align*}
  \vp^*
  &= {\rm TRI}(\vpi^{(1)}, \vpi^{(2)}, \vy^{1,L}, \vy^{1,R}, \vy^{2,L}, \vy^{2,R}) \\
  &= {\rm TRI}_{\rm rays}(\vp_{L,1}, \vr_{L,1}, \vp_{L,2}, \vr_{L,2}, \vp_{R,1}, \vr_{R,1}, \vp_{R,2}, \vr_{R,2}),
\end{align*}
where the ray origin $\vp_{C,j}(\vpi^{(j)})$ and bearing $\vr_{C,j} = \MR_C(\vq^{(j)}) \phi_C(\vy^{j,C})$ can be computed from \cref{eq:imu-to-camera}.
Then the Jacobian of the triangulated point $\vp^*$ with respect to $\vp^{(1)}$ can be computed using the chain rule as
\begin{equation}
  \frac{\partial \vp^*}{\partial \vp^{(1)}}
  = \frac{\partial {\rm TRI}_{\rm rays}}{\partial \vp_{L,1}}
  + \frac{\partial {\rm TRI}_{\rm rays}}{\partial \vp_{R,1}},
\end{equation}
because $\frac{\partial \vp_{C,1}}{\partial \vp^{(1)}} = I_3$ and $\frac{\partial \va}{\partial \vp^{(1)}} = \mbf{0}_3$ for all other arguments $\va$ of ${\rm TRI}_{\rm rays}$. The other blocks in the full Jacobian can be computed in a similar manner.

\vfill
\begin{figure}[!ht]
    \includegraphics[width=\linewidth]{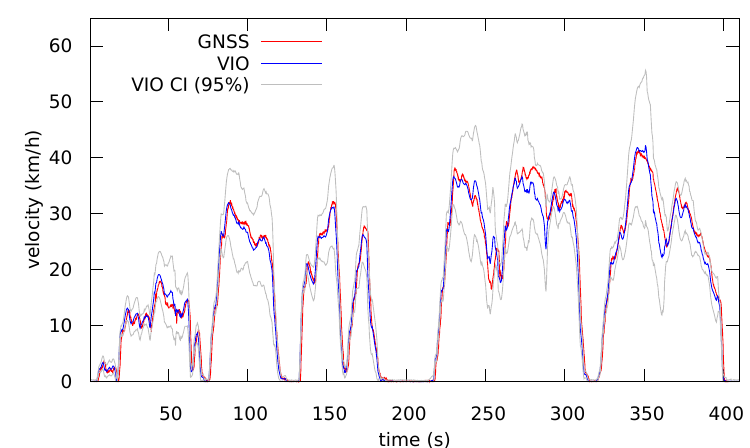}\vspace{0.75em}
    \caption{VIO velocity estimate for \cref{subfig:car-result}, \ourmethodname\ on ARKit}\vspace{8pt}
\end{figure}

\newpage
\paragraph{Quaternion update by angular velocity} If $\vomega = (\omega_x, \omega_y, \omega_z)$ represents a constant angular velocity, then a world-to-local quaternion $\vq = (q_w, q_x, q_y, q_z)^\T$ representing the orientation of a body transforms as
\begin{equation}
  \vq(t_0 + \Delta t) = \vectb{\Omega}[\vomega \Delta t] \vq(t_0)
\end{equation}
where
\begin{equation}
  \label{eq:quaternion-rotated-by-vector}
  \vectb{\Omega}[\vu] :=
  \exp\left[
  -\frac12
  \begin{pmatrix}
    0 & -u_x & -u_y & -u_z \\
    u_x & 0 & -u_z & u_y \\
    u_y & u_z & 0 & -u_x \\
    u_z & -u_y & u_x & 0
  \end{pmatrix}
  \right].
\end{equation}
Note that the matrix looks different if a local-to-world quaternion representation is used (\cf\ \cite{Titterton+Weston:2004}).

\vfill
\begin{figure}[!ht]
  \includegraphics[width=\linewidth]{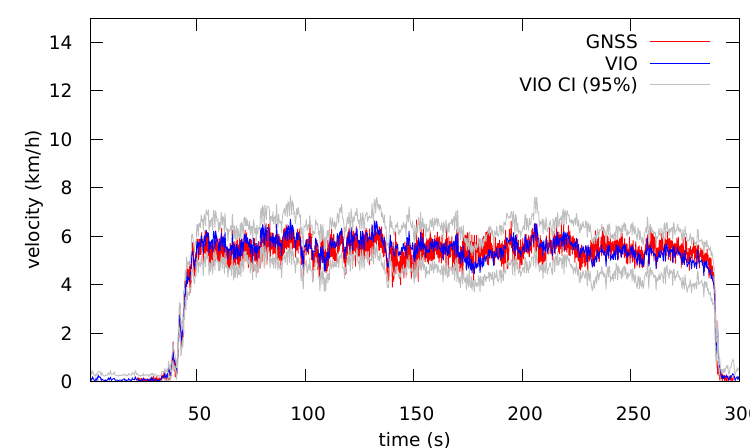}\vspace{0.75em}
  \caption{VIO velocity estimate for \cref{subfig:walking-result}, \ourmethodname\ on ARKit}\vspace{8pt}
\end{figure}

\begin{figure}[ht!]
  \centering
  \begin{subfigure}[b]{0.75\linewidth}
    \includegraphics[width=\linewidth]{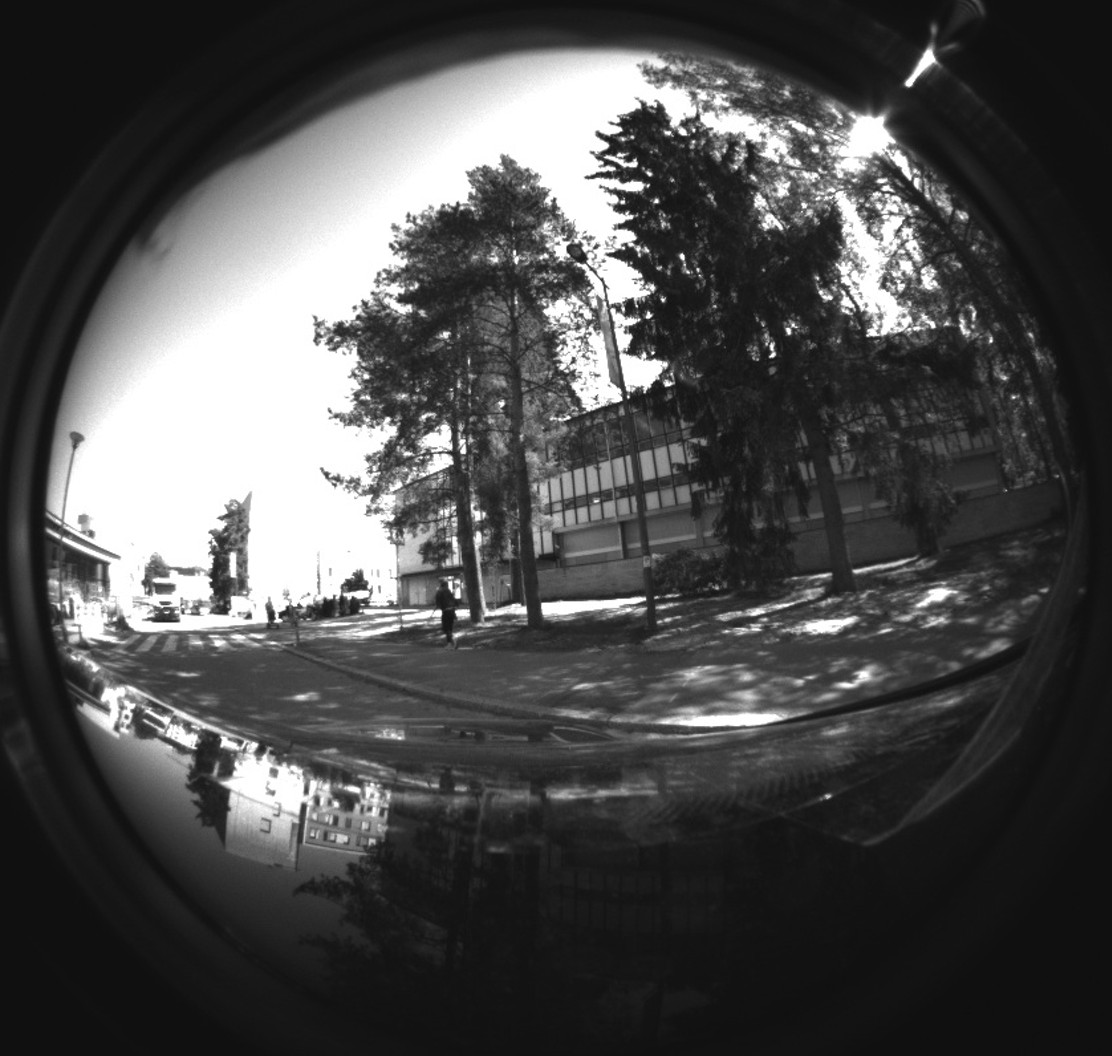}
    \caption{Intel RealSense T265 (left camera)}\vspace{8pt}
  \end{subfigure}
  \begin{subfigure}[b]{0.75\linewidth}
    \includegraphics[width=\linewidth]{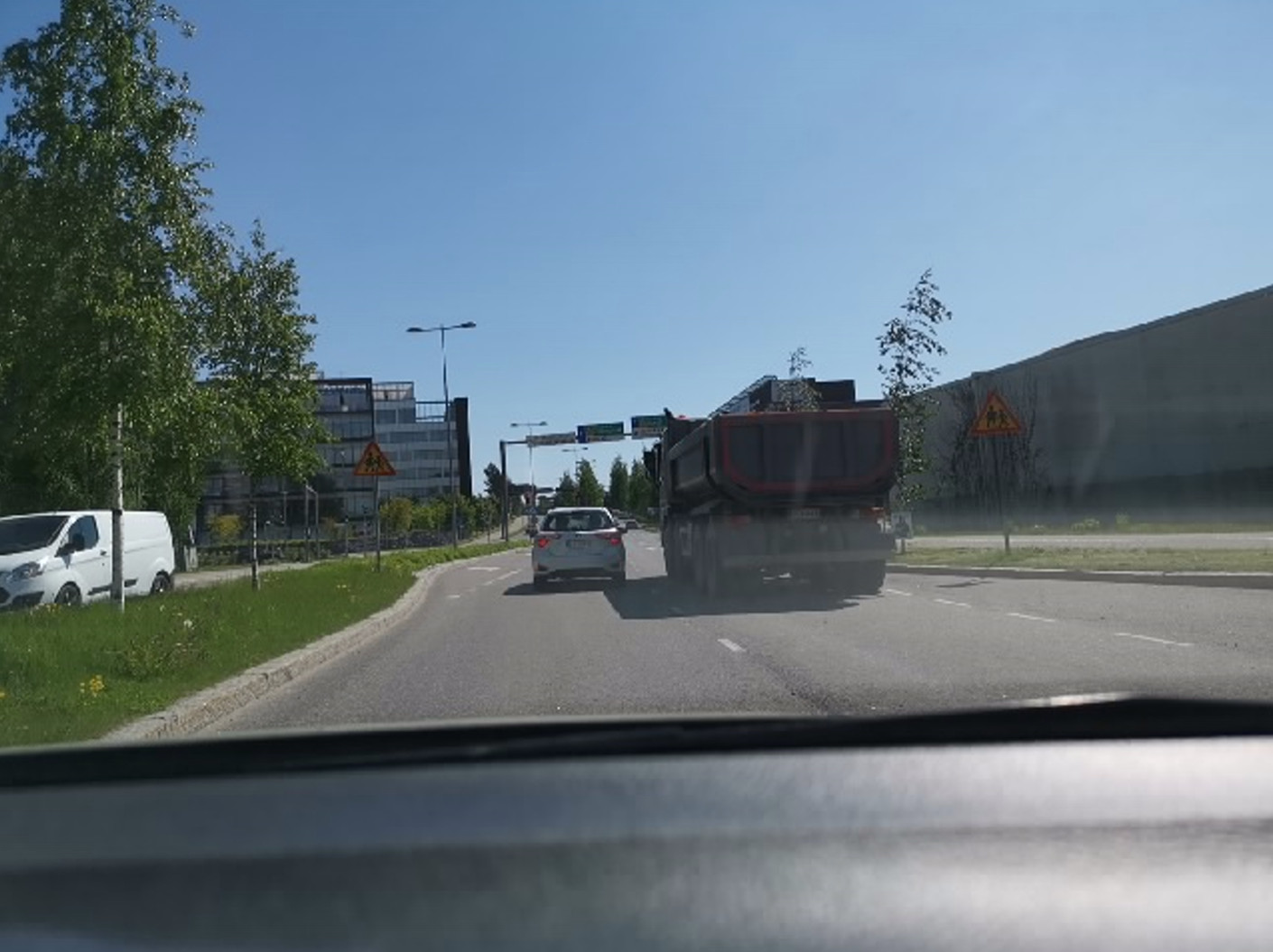}
    \caption{Huawei Mate 20 Pro (through ARCore)}\vspace{8pt}
  \end{subfigure}
  \begin{subfigure}[b]{0.75\linewidth}
    \includegraphics[width=\linewidth]{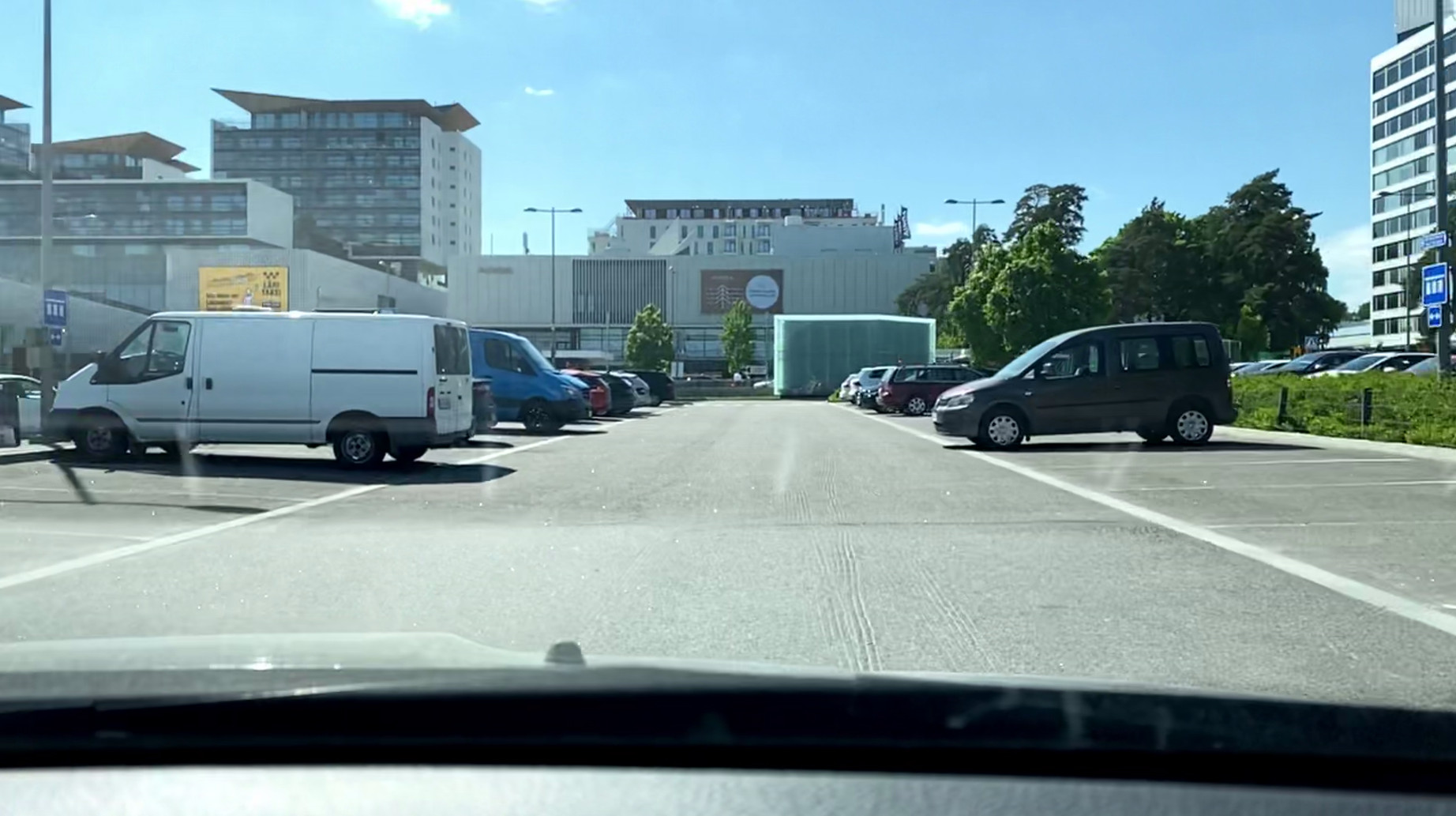}
    \caption{iPhone 11 Pro (through ARKit)}\vspace{8pt}
  \end{subfigure}
  \caption{Example camera views in the vehicular experiment \cref{subfig:car-result}. Reflections from the hood or windshield are visible in all images, and especially prominent in the RealSense fisheye camera.}
\end{figure}

\begin{figure}[ht!]%
  \centering%
  \begin{subfigure}[t]{0.9\linewidth}%
    \centering%
    \begin{tikzpicture}%
     \node[anchor=south west,inner sep=-3pt] (image) at (0,0) {\includegraphics[width=\linewidth]{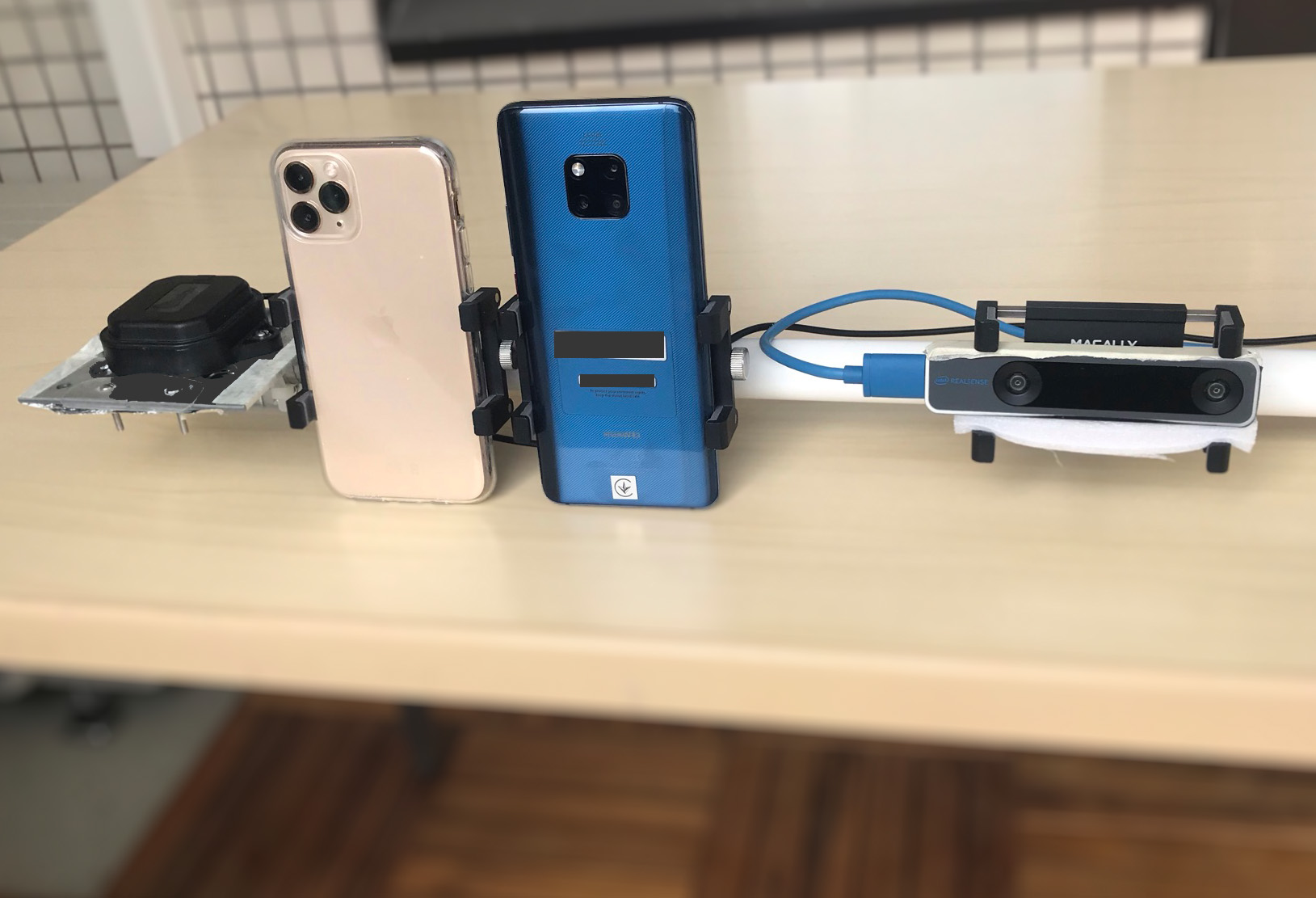}};%
     \begin{scope}[x={(image.south east)},y={(image.north west)}]%
         \fill [lightgray, rounded corners=2pt, opacity=.8] (0.05,0.07) rectangle (.95,.35);%
          \def\annotationcolor{black}%
          \node [anchor=west, \annotationcolor] (legendRS) at (0.55,0.27) {\small RealSense T265};
          \node [anchor=west, \annotationcolor] (legendRTK) at (0.1,0.15) {\small u-blox RTK-GNSS};
          \node [anchor=west, \annotationcolor] (legendIPhone) at (0.1,0.27) {\small iPhone 11 Pro};
          \node [anchor=west, \annotationcolor] (legendHuawei) at (0.55,0.15) {\small Huawei Mate 20 Pro};
         \draw [-latex, thick, \annotationcolor] (legendRS) to[out=0, in=270] (0.85,0.5);
         \draw [-latex, thick, \annotationcolor] (legendRTK) to[out=180, in=270] (0.13,0.55);
         \draw [-latex, thick, \annotationcolor] (legendIPhone) to[out=90, in=270] (0.3,0.43);
         \draw [-latex, thick, \annotationcolor] (legendHuawei) to[out=180, in=270] (0.5,0.43);
     \end{scope}
     \end{tikzpicture}%
    \caption{Devices, recorded as in \cref{fig:car-measurement-setup}\label{fig:walk-measurement-setup}}
 \end{subfigure}\\[0.5em]
 \begin{subfigure}[t]{0.9\linewidth}%
  \centering%
  \includegraphics[width=\linewidth]{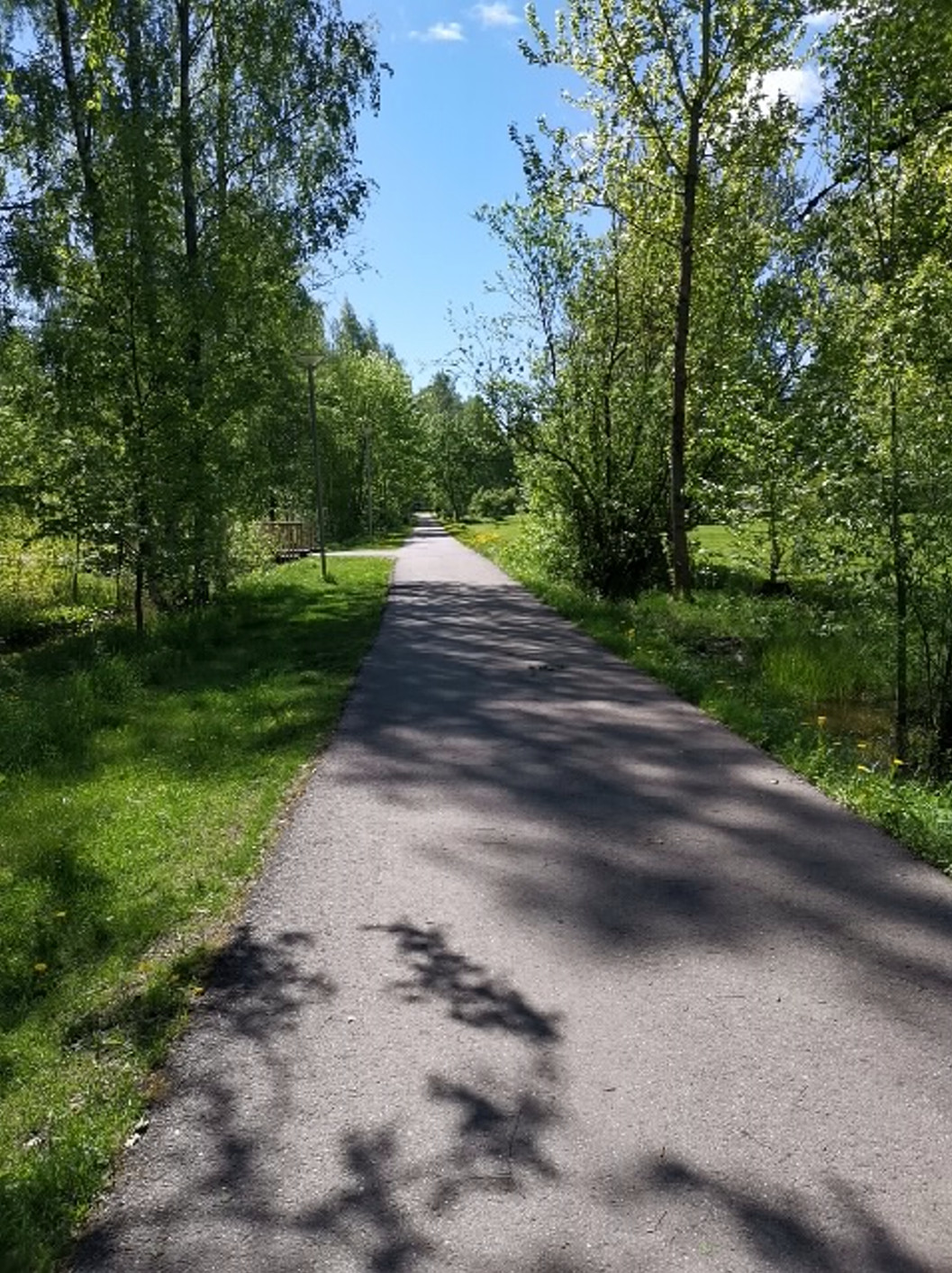}%
  \caption{Example camera view: Huawei Mate 20 Pro (through ARCore)}\vspace{8pt}
\end{subfigure}%
\caption{Walking experiment setup \cref{subfig:walking-result}}
\end{figure}

\setcounter{figure}{0}
\renewcommand{\thefigure}{B\arabic{figure}}%
\begin{figure*}
     \centering
     \begin{subfigure}[b]{\linewidth}
       \centering
       \input{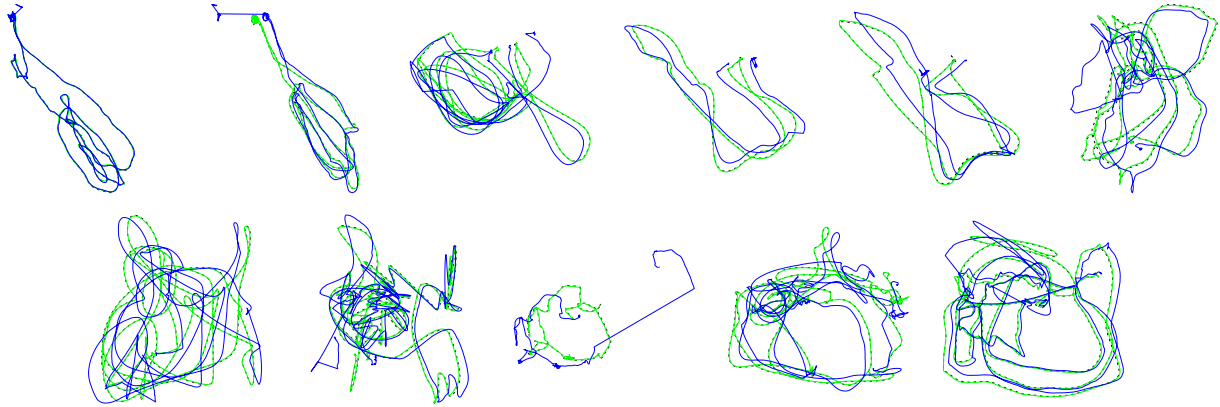}
       \caption{ORB-SLAM3. Due to successful loop closures, the method eventually recovers and is able to produce accurate post-processed trajectories.}
     \end{subfigure}\vspace{5pt}
     \begin{subfigure}[b]{\linewidth}
       \centering
       \input{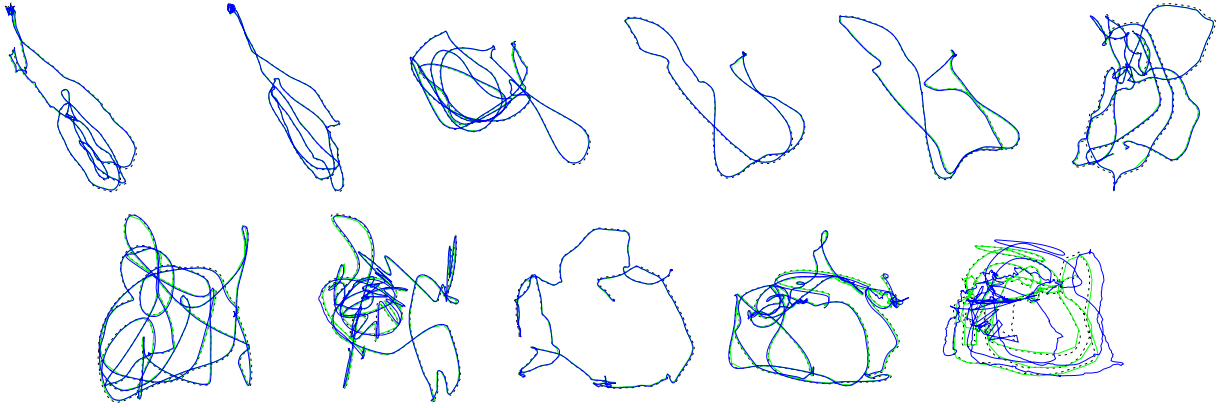}
       \caption{BASALT}
     \end{subfigure}\vspace{5pt}
     \begin{subfigure}[b]{\linewidth}
       \centering
       \input{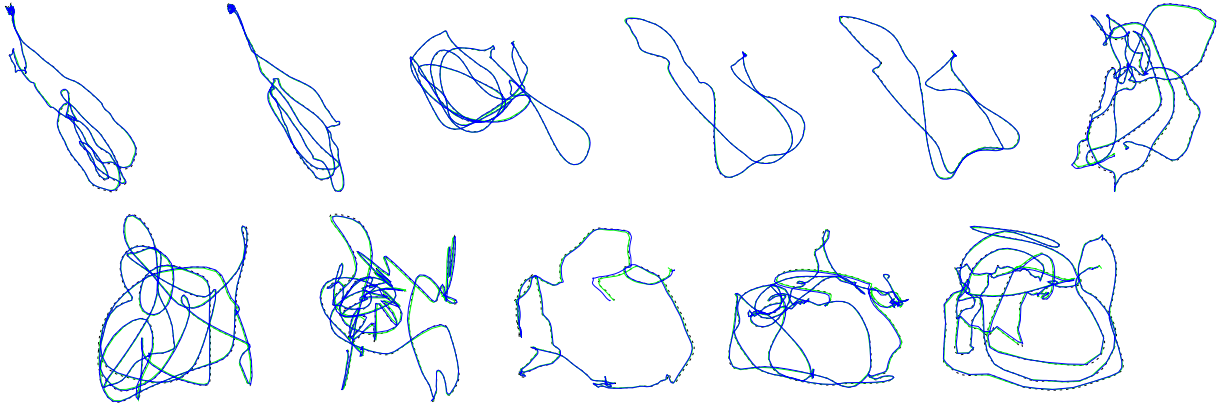}
       \caption{\ourmethodname\ (normal SLAM / post-processed SLAM)}
     \end{subfigure}\vspace{1em}
     \caption{Online (blue) and post-processed (green) trajectories in EuRoC MAV stereo mode, compared to ground truth (dashed) for three different methods. Our method and BASALT produce good results in both online and post-processed modes.
     \label{fig:euroc-online-ours-vs-orb-slam3}}
\end{figure*}

\clearpage

\widowpenalty=1000
\clubpenalty=1000
\section{Additional experiments}\label{sec:additional-experiments}
\setcounter{table}{0}
\renewcommand{\thetable}{B\arabic{table}}%
\subsection{Comparison to ORB-SLAM3 and BASALT}\label{sec:orb-slam-comparison}
\noindent
We processed the EuRoC dataset using the publicly available source code of ORB-SLAM\footnote{\url{https://github.com/UZ-SLAMLab/ORB_SLAM3} (V0.4)} and BASALT\footnote{\url{https://github.com/VladyslavUsenko/basalt-mirror} (June 7, 2021)} to compare execution times and reproduce the metrics reported in \cite{ORB-SLAM3, BASALT}. The ORB-SLAM3 example code was modified to output its intermediary results, namely the latest key frame pose after each input frame, without any changes to the actual algorithm. The results are presented in \cref{tbl:euroc-orbslam-basalt-run}.

All tests were performed on the same machine (the \emph{Ryzen} setup described in \cref{sec:experiments-euroc}) using two configurations: In the first, unrestricted configuration, the methods were allowed to utilize all 12 CPU cores in the system in parallel. BASALT was most efficiently parallelized and its processing time per frame was significantly lower than in the other methods. In the second configuration, we restricted the entire process (including input decoding) to use only 2 parallel CPU cores. In this mode, the processing times of the three methods were comparable.

The common results in \cref{tbl:euroc-orbslam-basalt-run} are similar to those reported in \cite{ORB-SLAM3, BASALT} (and reproduced in \cref{tbl:euroc-combined}). None of the methods (including ours) yield strictly equal results on different machines, which can explain the small remaining discrepancies. ORB-SLAM3 outputs also varied significantly between runs, but the online instability seen in \cref{fig:euroc-online-ours-vs-orb-slam3} was consistently observed.

\subsection{Ablation studies}\label{sec:additional-experiments-ablation}
\noindent
An ablation study equivalent to \cref{tbl:euroc-our-variants} for the TUM Room dataset is presented in \cref{tbl:tum-our-variants}. In the monocular case, the results are mixed: disabling individual novel features does not consistently improve the metrics, and the simple postprocessing actually degrades the SLAM results. However, most of our configurations, notably also \emph{Fast stereo VIO}, outperform all previous methods except ORB-SLAM3 (\cf\ \cref{tbl:tum-vi-room})

The SenseTime benchmark results in \cref{tbl:sensetime-our-variants} are more consistent and similar to \cref{tbl:euroc-our-variants}: the novel features are all beneficial and the baseline PIVO implementation is not stable. However, our simple post-processing method is not able to improve the results compared to the online case.

\cref{tbl:euroc-parameter-variations} studies the effect of varying the parameters presented in \cref{tbl:parameters} individually. Deviations from the selected parameters caused degraded metrics, except increasing $n_{\rm BA}$ improves the baseline results. However, this larger bundle adjustment problem is too heavy for the real-time use case and therefore we only use it in the post-processed setting.

\begin{table*}
  \caption{EuRoC computational times and RMSE in different methods (stereo SLAM). Also shown in \cref{fig:euroc-online-ours-vs-orb-slam3}.}
  \vspace{2pt}
  \label{tbl:euroc-orbslam-basalt-run}
  \centering
  \ourtablefontsize
  \renewcommand{\arraystretch}{1.1}
  \begin{tabular}{>{\raggedright}P{3.5em}|p{6em}P{2.1em}P{2.1em}P{2.1em}P{2.1em}P{2.1em}P{2.1em}P{2.1em}P{2.1em}P{2.1em}P{2.1em}P{2.1em}|P{2.1em}|P{3.78em}P{3.78em}}
 &  &  &  &  &  &  &  &  &  &  &  &  &  & \multicolumn{2}{m{9.072em}}{Ryzen frame time (ms)}\\
 & Method & MH01 & MH02 & MH03 & MH04 & MH05 & V101 & V102 & V103 & V201 & V202 & V203 & Mean & all CPUs & 2 CPUs\\
\toprule
online & Ours\textsuperscript{(1)} & 0.088 & 0.080 & 0.038 & 0.071 & 0.108 & 0.044 & 0.035 & 0.040 & 0.075 & 0.041 & 0.052 & 0.061 & 32 & 47\\
 & ORB-SLAM3 & 0.094 & 1.229 & 1.124 & 1.887 & 2.177 & 0.698 & 2.036 & 0.529 & 3.488 & 1.498 & 0.445 & 1.382 & 56 & 78\\
 & BASALT & 0.080 & 0.052 & 0.078 & 0.106 & 0.120 & 0.045 & 0.058 & 0.088 & 0.035 & 0.073 & 0.897 & 0.148 & 5 & 36\\
\cline{1-16}
post-pr. & Ours\textsuperscript{(3)} & 0.048 & 0.028 & 0.037 & 0.056 & 0.066 & 0.038 & 0.035 & 0.037 & 0.031 & 0.029 & 0.044 & 0.041 & 52 & 95\\
 & ORB-SLAM3 & 0.033 & 0.030 & 0.031 & 0.056 & 0.100 & 0.036 & 0.014 & 0.025 & 0.037 & 0.016 & 0.019 & 0.036 & 56 & 78\\
 & BASALT & 0.085 & 0.065 & 0.056 & 0.105 & 0.099 & 0.046 & 0.033 & 0.035 & 0.041 & 0.028 & 0.175 & 0.070 & 14 & 66\\
\cline{1-16}\end{tabular}

\end{table*}

\begin{table}
  \caption{Different configurations of \ourmethodname\ on the TUM Room dataset (\cf\ \cref{tbl:euroc-our-variants} and \cref{tbl:tum-vi-room}).}
  \vspace{2pt}
  \label{tbl:tum-our-variants}
  \centering
  \ourtablefontsize
  \setlength{\tabcolsep}{5.2pt}
  \renewcommand{\arraystretch}{1.1}
  \begin{tabular}{>{\raggedright}P{0.5em}|P{0.5em}|p{6.1em}P{1.6em}P{1.6em}P{1.6em}P{1.6em}P{1.6em}P{1.6em}|P{1.6em}}
 &  & Method & R1 & R2 & R3 & R4 & R5 & R6 & Mean\\
\toprule
 &  & Normal SLAM & 0.016 & 0.013 & 0.011 & 0.013 & 0.02 & 0.01 & 0.014\\
\cline{3-10}
\multirow{14}{*}{\rotatebox{90}{online}} & \multirow{4}{*}{\rotatebox{90}{stereo}} & Normal VIO & 0.05 & 0.053 & 0.041 & 0.042 & 0.082 & 0.033 & 0.050\\
 &  & $\smallsetminus$ \cref{eq:anti-track-reuse-criterion} & 0.072 & 0.052 & 0.037 & 0.042 & 0.11 & 0.058 & 0.062\\
\cline{3-10}
 &  & Fast VIO & 0.075 & 0.064 & 0.074 & 0.041 & 0.07 & 0.037 & 0.060\\
 &  & $\smallsetminus$ \cref{eq:fifo-and-towers-of-hanoi-scheme} & 0.09 & 0.11 & 0.055 & 0.052 & 0.065 & 0.083 & 0.076\\
\cline{2-10}
 & \multirow{10}{*}{\rotatebox{90}{mono}} & Normal SLAM & 0.02 & 0.02 & 0.17 & 0.018 & 0.019 & 0.017 & 0.044\\
\cline{3-10}
 &  & Normal VIO & 0.08 & 0.06 & 0.17 & 0.036 & 0.079 & 0.06 & 0.080\\
 &  & $\smallsetminus$ RANSAC & 0.065 & 0.072 & 0.092 & 0.058 & 0.06 & 0.049 & 0.066\\
 &  & $\smallsetminus$ \cref{eq:imu-bias-propagation} & 0.089 & 0.062 & 0.23 & 0.057 & 0.094 & 0.07 & 0.101\\
 &  & $\smallsetminus$ \cref{eq:longer-than-median-tracks} & 0.09 & 0.066 & 0.21 & 0.05 & 0.069 & 0.049 & 0.089\\
 &  & $\smallsetminus$ \cref{sec:stationarity-detection} & 0.087 & 0.06 & 0.14 & 0.046 & 0.079 & 0.06 & 0.078\\
 &  & $\smallsetminus$ \cref{eq:anti-track-reuse-criterion} & 0.083 & 0.083 & 0.081 & 0.07 & 0.066 & 0.068 & 0.075\\
 &  & PIVO baseline & 0.075 & 0.077 & 0.11 & 0.051 & 0.14 & 0.071 & 0.088\\
\cline{3-10}
 &  & Fast VIO & 0.086 & 0.061 & 0.066 & 0.077 & 0.061 & 0.07 & 0.070\\
 &  & $\smallsetminus$ \cref{eq:fifo-and-towers-of-hanoi-scheme} & 0.09 & 0.062 & 0.12 & 0.082 & 0.08 & 0.051 & 0.082\\
\cline{1-10}
\multicolumn{2}{P{2.2em}|}{\multirow{2}{*}{post-pr.}} & Stereo SLAM & 0.042 & 0.041 & 0.028 & 0.025 & 0.061 & 0.02 & 0.036\\
\multicolumn{2}{P{2.2em}|}{} & Mono SLAM & 0.039 & 0.033 & 0.16 & 0.032 & 0.039 & 0.026 & 0.055\\
\cline{1-10}\end{tabular}

\end{table}
\begin{table}
  \caption{Different configurations of \ourmethodname\ in the SenseTime benchmark (\cf\ \cref{tbl:euroc-our-variants} and \cref{tbl:sensetime}).}
  \vspace{2pt}
  \setlength{\tabcolsep}{5.6pt}
  \label{tbl:sensetime-our-variants}
  \centering
  \ourtablefontsize
  \renewcommand{\arraystretch}{1.1}
  \begin{tabular}{>{\raggedright}P{0.5em}|p{6.2em}|P{1em}P{1em}P{1em}P{1em}P{1em}P{1em}P{1em}P{1em}|P{1em}}
 & Method & A0 & A1 & A2 & A3 & A4 & A5 & A6 & A7 & Mean\\
\cline{1-11}
 & Normal SLAM & 49.9 & 30 & 36 & 22.2 & 19.6 & 37.8 & 29.3 & 17.3 & 30.3\\
\cline{2-11}
\multirow{9}{*}{\rotatebox{90}{online}} & Normal VIO & 63.5 & 53.1 & 53.9 & 28.1 & 26.3 & 75.6 & 29.7 & 26.6 & 44.6\\
 & $\smallsetminus$ RANSAC & 73.1 & 109 & 59 & 27.6 & 24.2 & 65.8 & 30 & 22 & 51.3\\
 & $\smallsetminus$ \cref{eq:imu-bias-propagation} & 171 & 272 & 208 & 40.8 & 83.6 & 170 & 116 & 120 & 148\\
 & $\smallsetminus$ \cref{eq:longer-than-median-tracks} & 69.5 & 70.9 & 47.8 & 33.9 & 31.4 & 74.9 & 32.2 & 43.1 & 50.5\\
 & $\smallsetminus$ \cref{sec:stationarity-detection} & 87.7 & 764 & 149 & 75.6 & 158 & 351 & 310 & 328 & 278\\
 & $\smallsetminus$ \cref{eq:anti-track-reuse-criterion} & 75.2 & 46.3 & 53 & 27.7 & 34 & 72.9 & 31.2 & 27.6 & 46\\
 & PIVO baseline & 166 & 1150 & 225 & 219 & 242 & 472 & 109 & 239 & 353\\
\cline{2-11}
 & Fast VIO & 47.2 & 59.6 & 43.1 & 25.8 & 46 & 60.5 & 30.6 & 40 & 44.1\\
 & $\smallsetminus$ \cref{eq:fifo-and-towers-of-hanoi-scheme} & 80 & 73.4 & 89.1 & 26.8 & 50 & 119 & 38 & 50.2 & 65.8\\
\cline{1-11}
\multicolumn{2}{P{8em}|}{Post-pr. SLAM} & 63.9 & 28.4 & 28.5 & 23.2 & 42.7 & 23.3 & 22.2 & 18.1 & 31.3\\
\cline{1-11}\end{tabular}

\end{table}
\begin{table}
  \caption{Effect of individual parameters in \cref{tbl:parameters} on the mean RMSE and frame time in EuRoC. The baseline is \emph{Normal SLAM}.}
  \vspace{4pt}
  \label{tbl:euroc-parameter-variations}
  \centering
  \ourtablefontsize
  \renewcommand{\arraystretch}{1}
  \begin{tabular}{>{\raggedright}p{8em}|P{2.16em}|P{2.16em}|l}
Altered parameter & Value & RMSE & Frame time (ms)\\
\toprule
baseline\textsuperscript{(1)} &  & 0.061 & 35\\
\cline{1-4}
feature detector & FAST & 0.067 & 33\\
subpix. adjustment & off & 0.065 & 33\\
max. features & 70 & 0.066 & 19\\
max. features & 100 & 0.065 & 23\\
max. features & 300 & 0.061 & 48\\
max. itr. & 8 & 0.064 & 35\\
max. itr. & 40 & 0.066 & 36\\
window size & 13 & 0.062 & 34\\
window size & 51 & 0.07 & 37\\
$n_{\rm a}$ & 30 & 0.062 & 43\\
$n_{\rm {target}}$ & 5 & 0.063 & 33\\
$n_{\rm {target}}$ & 10 & 0.064 & 37\\
$n_{\rm {target}}$ & 30 & 0.061 & 34\\
$n_{\rm {FIFO}}$ & 20 & 0.068 & 36\\
$n_{\rm {FIFO}}$ & 14 & 0.31 & 35\\
$n_{\rm {BA}}$ & 50 & 0.053 & 42\\
$n_{\rm {BA}}$ & 100 & 0.055 & 79\\
$n_{\rm {matching}}$ & 35 & 0.06 & 36\\
$n_{\rm {matching}}$ & 50 & 0.06 & 36\\
\cline{1-4}\end{tabular}

\end{table}

\subsection{Vehicular}\label{sec:additional-experiments-vehicular}
\begin{figure}
  \begin{subfigure}[b]{\linewidth}
    \includegraphics[width=\linewidth]{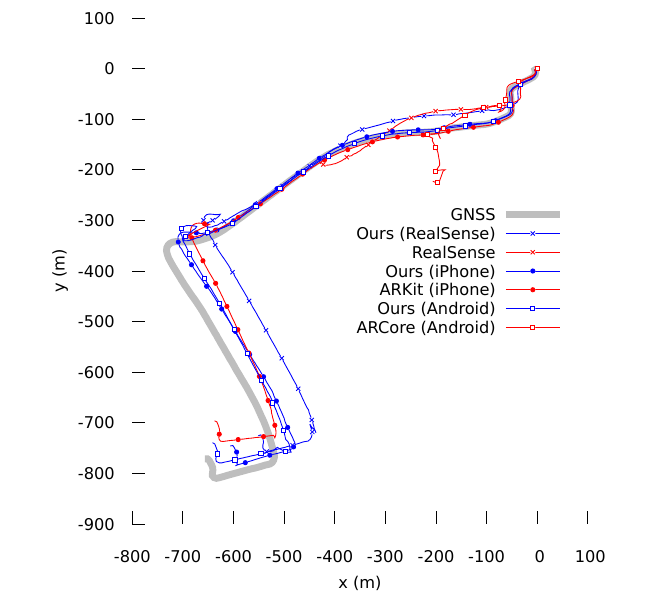}
    \caption{Position}\vspace{8pt}
  \end{subfigure}
  \begin{subfigure}[b]{\linewidth}
    \includegraphics[width=\linewidth]{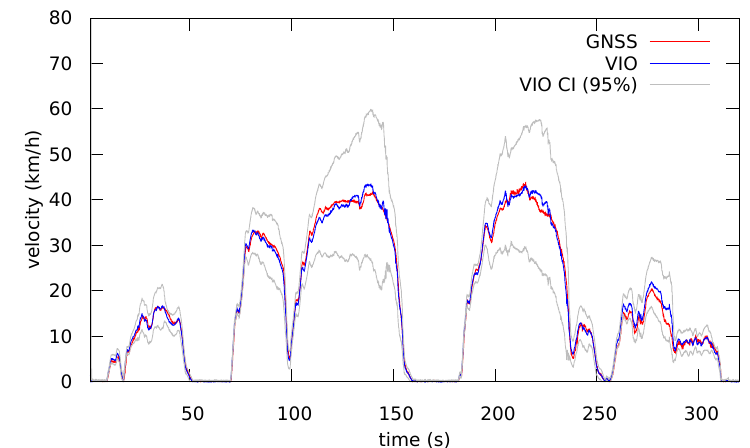}
    \caption{VIO velocity estimate, \ourmethodname\ on ARKit}\vspace{8pt}
  \end{subfigure}
  \caption{Vehicular experiment 2\label{fig:car-realsense-01}, using the setup in \cref{fig:car-measurement-setup}}
\end{figure}

\begin{figure}
  \begin{subfigure}[b]{0.9\linewidth}
    \includegraphics[width=\linewidth]{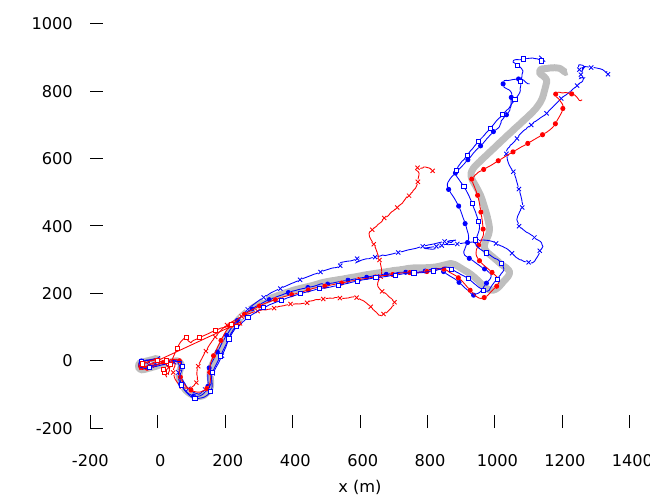}
    \caption{Position}\vspace{8pt}
  \end{subfigure}
  \begin{subfigure}[b]{0.9\linewidth}
    \includegraphics[width=\linewidth]{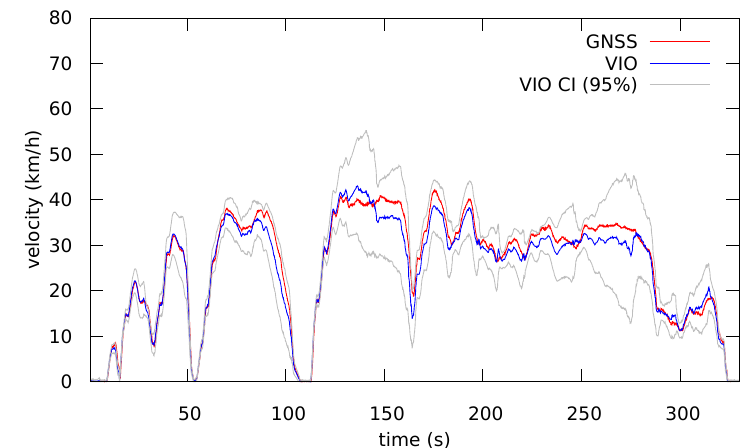}
    \caption{VIO velocity estimate, \ourmethodname\ on ARKit}\vspace{8pt}
  \end{subfigure}
  \caption{Vehicular experiment 3, using the setup in \cref{fig:car-measurement-setup}\label{fig:car-realsense-03}}
\end{figure}

\begin{figure*}
  \begin{subfigure}[t]{0.45\linewidth}
       \centering
       \begin{tikzpicture}
        \node[anchor=south west,inner sep=0] (image) at (0,0) {\includegraphics[width=\linewidth]{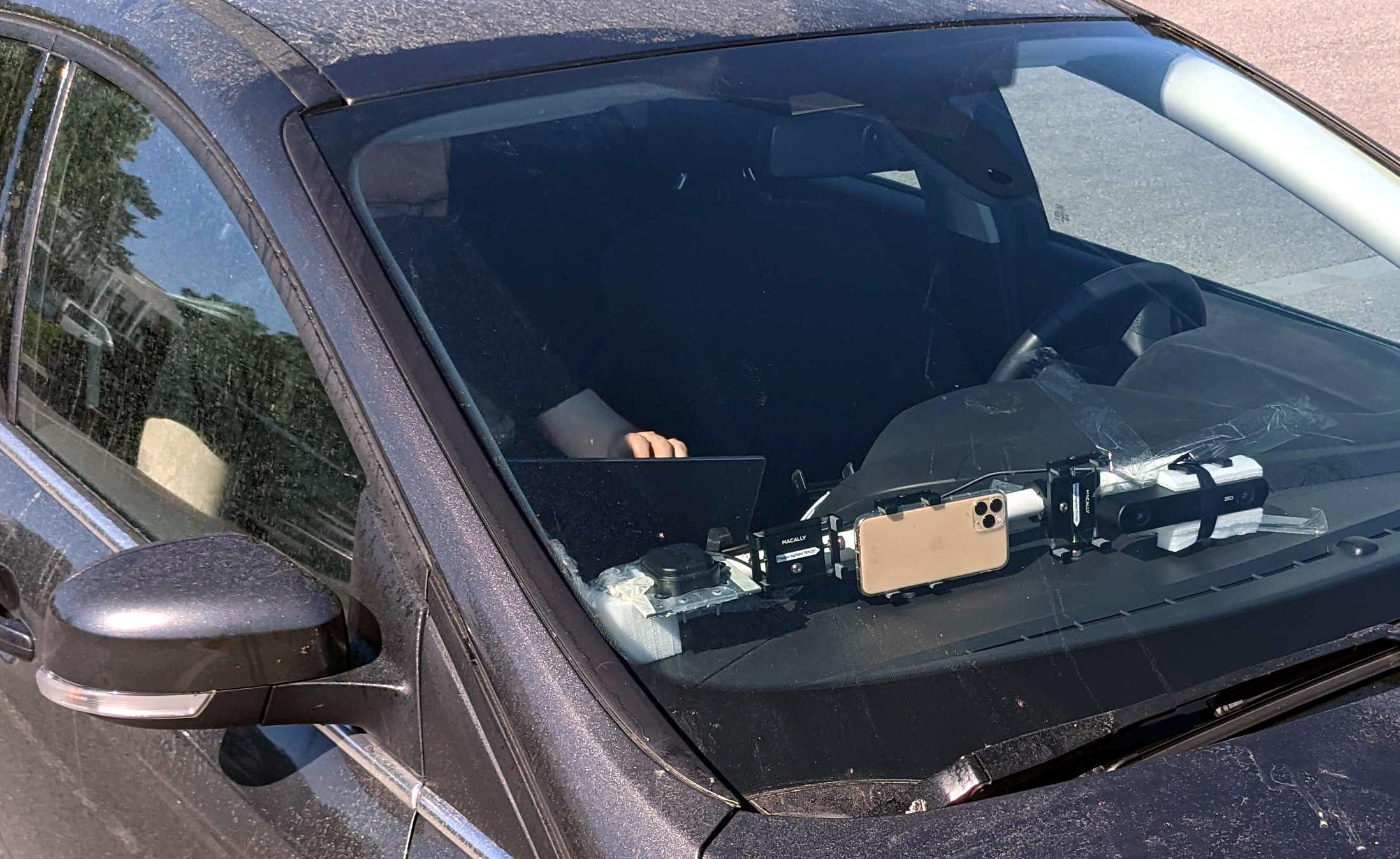}};
        \begin{scope}[x={(image.south east)},y={(image.north west)}]
            \fill [gray, rounded corners=2pt, opacity=.8] (0.61,0.72) rectangle (1,1);
             \def\annotationcolor{yellow}
             \node [anchor=west, \annotationcolor] (legendRTK) at (0.63,0.93) {\small u-blox RTK-GNSS};
             \node [anchor=west, \annotationcolor] (legendIPhone) at (0.63,0.86) {\small iPhone 11 Pro};
             \node [anchor=west, \annotationcolor] (legendHuawei) at (0.63,0.79) {\small Zed 2};
            \draw [-latex, thick, \annotationcolor] (legendRTK) to[out=180, in=90] (0.48,0.4);
            \draw [-latex, thick, \annotationcolor] (legendIPhone) to[out=180, in=120] (0.65,0.45);
            \draw [-latex, thick, \annotationcolor] (legendHuawei) to[out=270, in=110] (0.83,0.5);
        \end{scope}
       \end{tikzpicture}\vspace{3pt}
       \caption{Second car experiment setup: GNSS is used as ground truth. The iPhone records ARKit and its input data simultaneously. ZED 2 records camera (stereo rolling shutter at 60FPS) and IMU data.\label{fig:car-measurement-setup2}}
  \end{subfigure}\vspace{0.8em}
  \hfill
  \begin{subfigure}[t]{0.49\linewidth}
    \includegraphics[width=\linewidth]{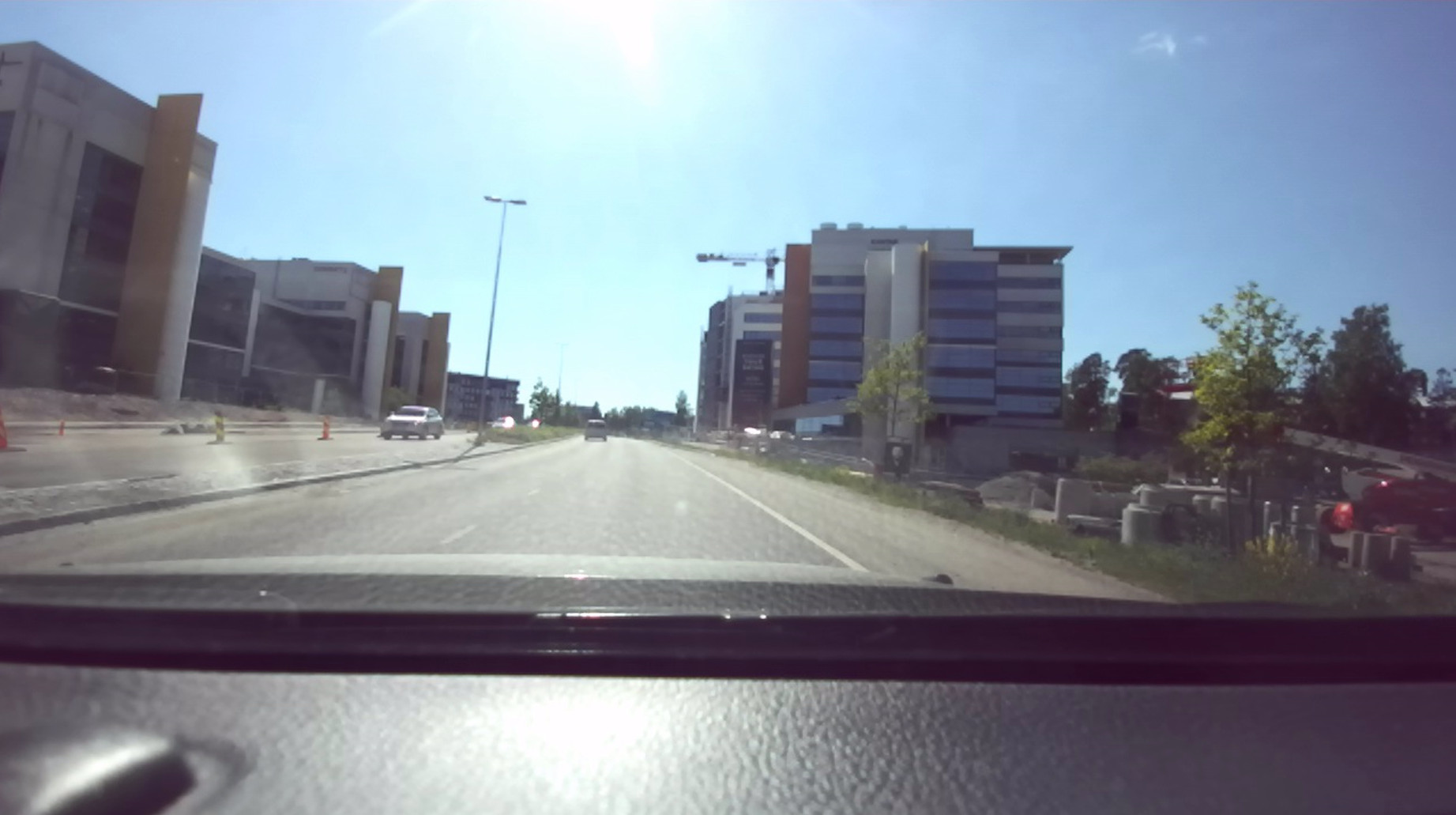}
    \caption{Example ZED 2 left camera view corresponding to (\subref{subfig:car-22-pos})}\vspace{8pt}
  \end{subfigure}

  \hfill\begin{subfigure}[b]{0.49\linewidth}
    \includegraphics{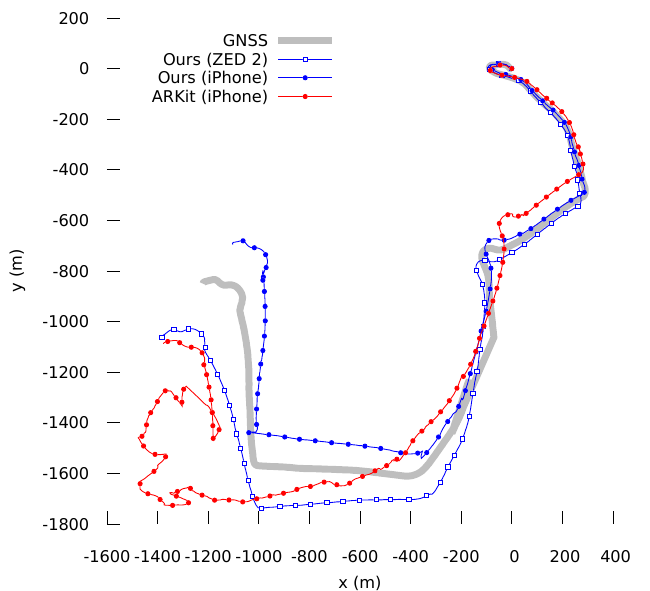}
    \caption{Vehicular experiment 5\label{subfig:car-22-pos}}\vspace{8pt}
  \end{subfigure}
  \begin{subfigure}[b]{0.49\linewidth}
    \includegraphics[width=\linewidth]{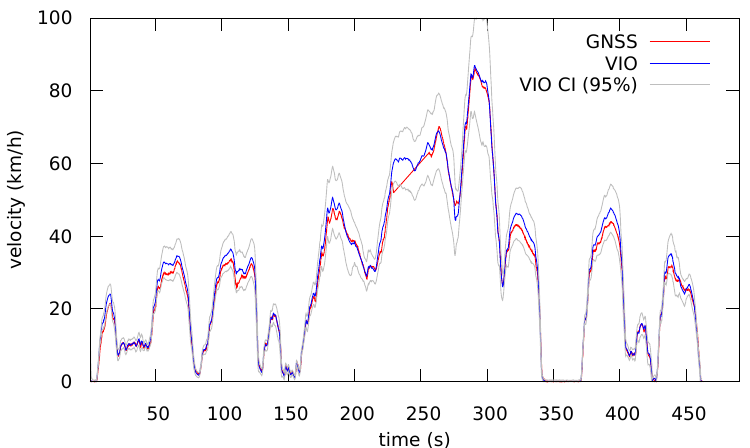}
    \caption{VIO velocity estimate for (\subref{subfig:car-22-pos}), \ourmethodname\ on ZED 2. A GNSS outage is visible as a straight line segment near the 250 seconds mark.}\vspace{8pt}
  \end{subfigure}

  \hfill\begin{subfigure}[b]{0.49\linewidth}
    \includegraphics{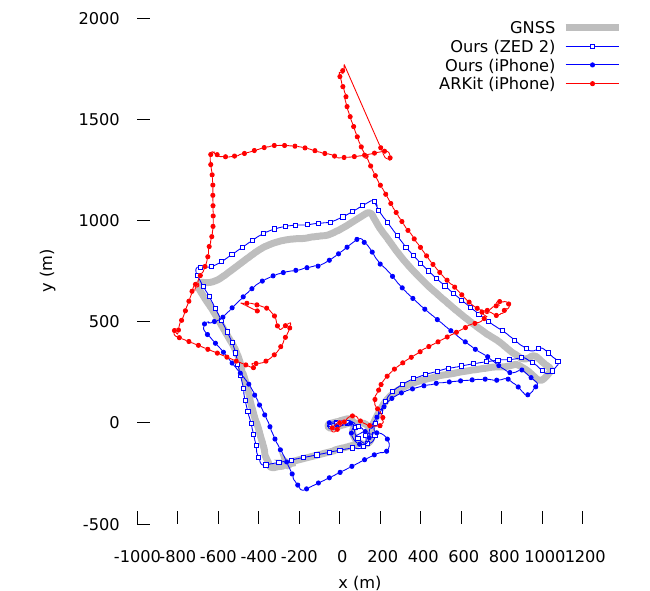}
    \caption{Vehicular experiment 6\label{subfig:car-24-pos}}\vspace{8pt}
  \end{subfigure}
  \begin{subfigure}[b]{0.49\linewidth}
    \includegraphics[width=\linewidth]{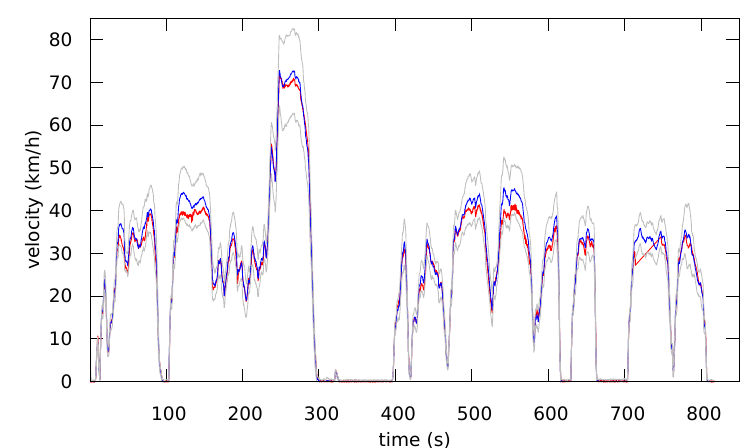}
    \caption{VIO velocity estimate for (\subref{subfig:car-24-pos}), \ourmethodname\ on ZED 2. A GNSS outage is visible after 700 seconds.}\vspace{8pt}
  \end{subfigure}
  \caption{Additional vehicular experiments with higher velocities ($\sim$ 80km/h), in which ARKit also fails. In this case, \ourmethodname\ performs better on the same data.}
\end{figure*}

\noindent%
This section includes additional vehicular experiments (\cref{fig:car-realsense-01} and \cref{fig:car-realsense-03}) using the setup shown in \cref{fig:car-measurement-setup}, as well as results from a slightly modified setup (shown in \cref{fig:car-measurement-setup2}), where we have added ZED 2 as a new device.

\begin{figure}
  \centering
  \includegraphics[width=0.9\linewidth]{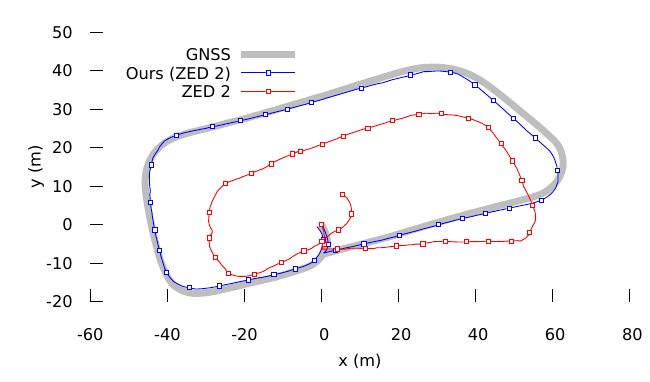}
  \caption{Vehicular experiment 4. A slow drive around a parking lot, recording the setup shown in \cref{fig:car-measurement-setup2}. Unlike the following experiments with ZED 2, the proprietary tracking output from ZED 2 is compared to \ourmethodname\ using the same input data.\label{fig:car-zed-01}}
\end{figure}

ZED 2 also has a proprietary VISLAM capability, but it did not perform well in the vehicular test cases (see \cref{fig:car-zed-01} for an example) and we omitted it in the other sequences to avoid frame drop issues experienced when recording ZED 2 input and tracking output data simultaneously. The ZED 2 camera data was recorded at 60FPS but utilized at 30FPS.

We used the \emph{normal VIO} mode (see \cref{tbl:parameters}) for all vehicular experiments. Stereo mode was used with both stereo camera devices.

\clearpage
\clearpage
\section{Supplementary videos}
\setcounter{figure}{0}
\renewcommand{\thefigure}{C\arabic{figure}}%
\noindent%

\begin{enumerate}[label=(\alph*)]
\item {\tt \textbf{euroc-mh-05-difficult\_fast-VIO}} \\ (\href{https://youtu.be/ou1DrtjPx1Q}{\tt\textbf{https://youtu.be/ou1DrtjPx1Q}}) A screen recording from a laptop running \ourmethodname\ in \emph{fast VIO} mode (\cf\ \cref{tbl:parameters}) on the EuRoC MAV sequence MH-05. The final trajectory also appears in \cref{fig:teaser-euroc-uncertainty}.

The visual tracking and update status are visualized on the left and right camera frames, similarly to \cref{fig:euroc-tracker-example} but with different colors: reprojections are in white, successfully updated tracks in black and failed tracks in blue. The lower part of the video shows the online track (x and y coordinates) in blue and ground truth in orange. Triangulated points are shown as small black circles. The online track is automatically rotated to optimally match the ground truth, after approximately 10 seconds.

\item {\tt \textbf{euroc-v1-02-medium\_normal-SLAM}} \\ (\href{https://youtu.be/7j1rYoD_pPc}{\tt\textbf{https://youtu.be/7j1rYoD\_pPc}}) Similar to the previous video, but \ourmethodname\ is running in the \emph{normal SLAM} mode on the sequence V1-02. The triangulated SLAM map points in the current local map are shown as yellow in the lower right subimage, and their reprojections as orange on the (left) camera image. The LK-tracked features (\cf\ \cref{algo:hybrid-vio-slam}) that correspond to SLAM map points are shown as yellow circles on the camera image.

The left part of the video shows the SLAM map. Key frame camera poses are shown in light blue. In the beginning, the triangulated points in the local map are shown in red. At time 00:27, the colors are changed to show the observation direction of the map point. The \emph{covisibility graph} is shown first at 00:15, in a yellow-green color. We consider a pair of key frames adjacent in this graph if they observe at least $N_{\rm neigh} = 10$ common map points.

\item {\tt \textbf{vehicular-experiment-6}} \\ (\href{https://youtu.be/iVNicL_S14Y}{\tt\textbf{https://youtu.be/iVNicL\_S14Y}}) Visualizes the vehicular experiment in \cref{subfig:car-24-pos} on a map (\ourmethodname\ on ZED 2). The VIO trajectory is aligned using a fixed angle and offset. The \emph{local VIO trajectory} is formed using the pose trails (\cf\ \cref{sec:vio-state}) in the VIO state. The traffic light stops are automatically cut from the video (based on the VIO velocity estimate). Despite generally good RTK-GNSS coverage in the area, the sequence includes a GNSS outage in a tunnel, starting at time 02:32 in the video.
\end{enumerate}

\begin{figure}[ht!]
  \centering
  \begin{subfigure}[b]{0.95\linewidth}
    \includegraphics[width=\linewidth]{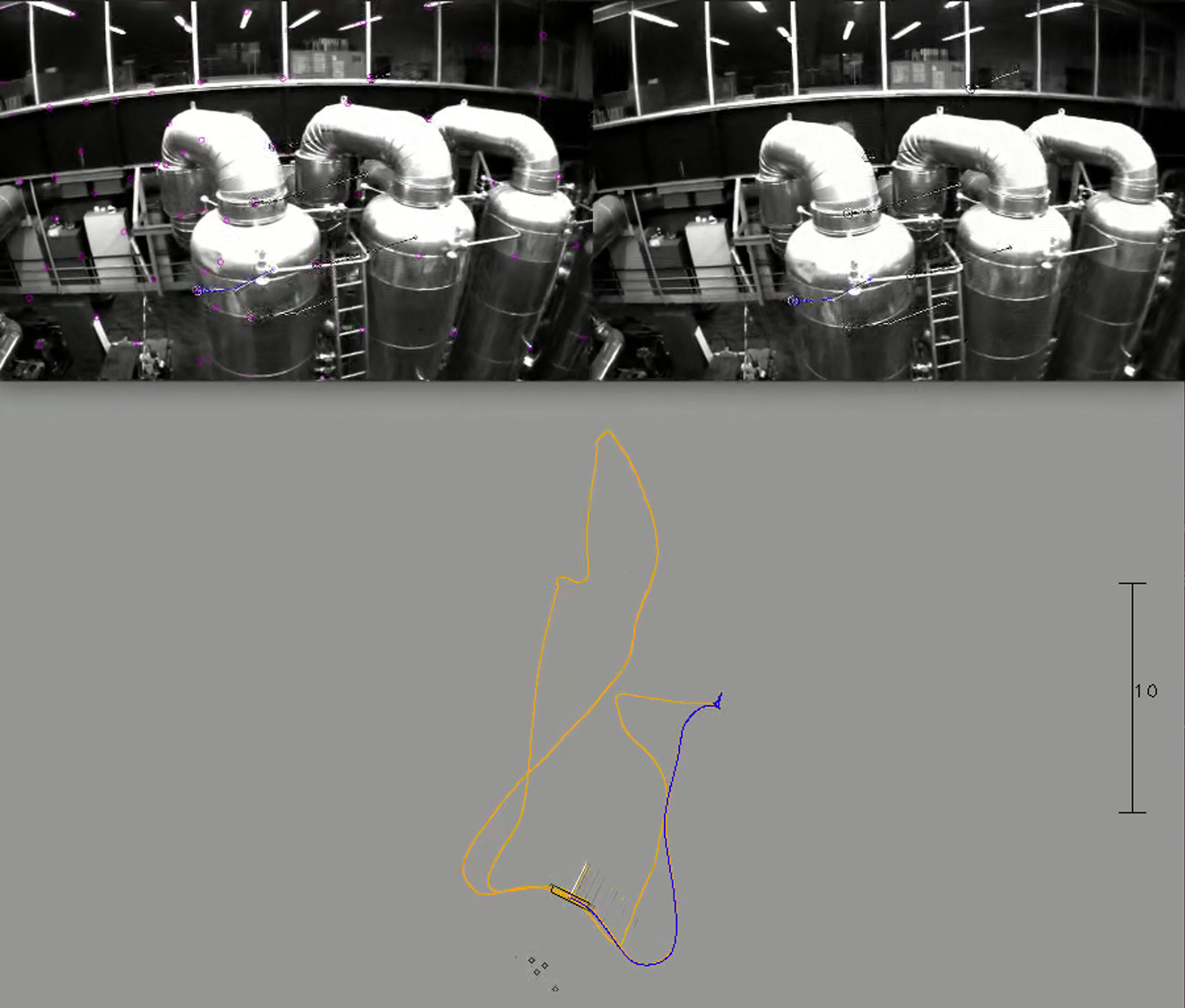}
    \caption{{\tt {euroc-mh-05-difficult\_fast-VIO}}}\vspace{8pt}
  \end{subfigure}
  \begin{subfigure}[b]{0.95\linewidth}
    \includegraphics[width=\linewidth]{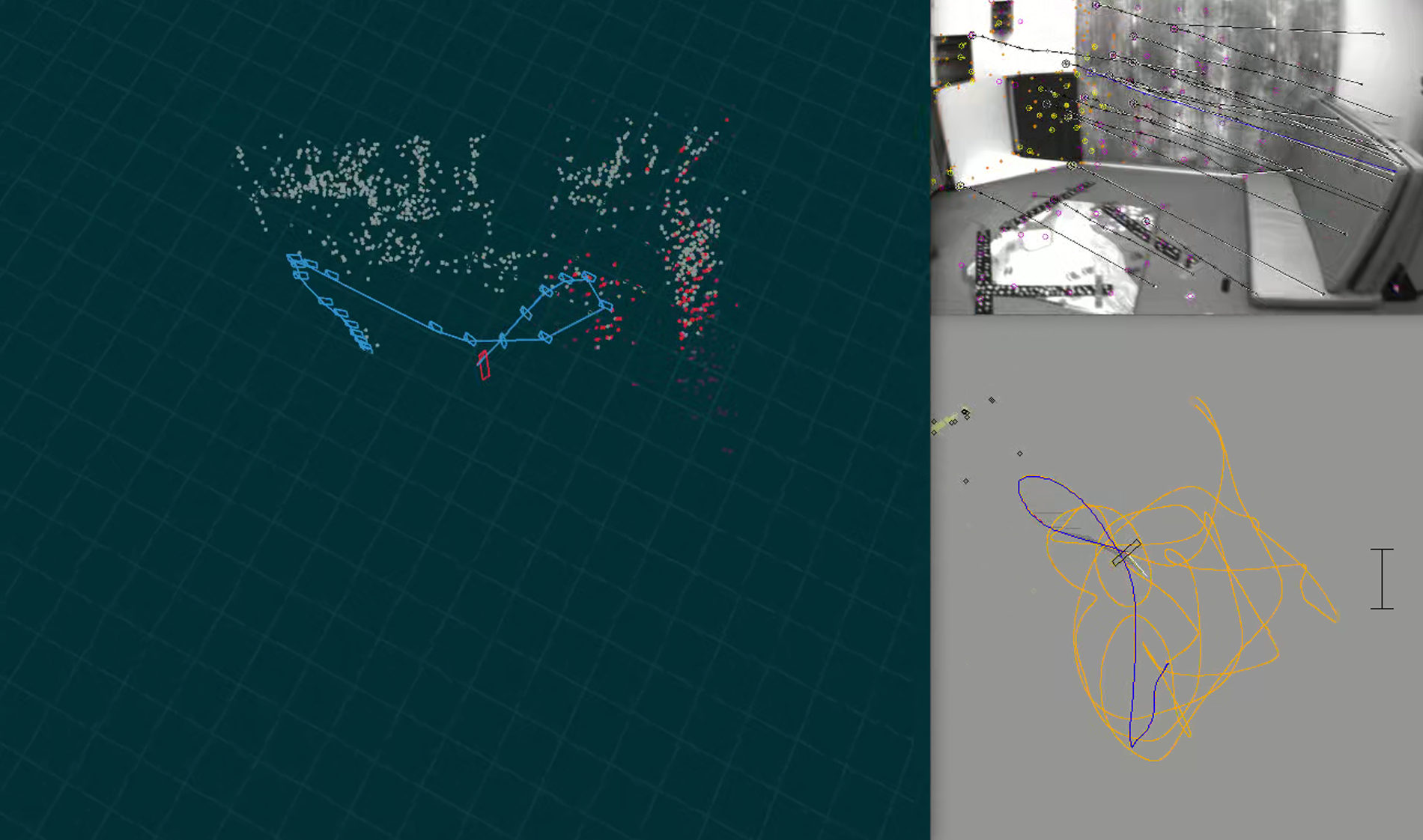}
    \caption{{\tt {euroc-v1-02-medium\_normal-SLAM}}}\vspace{8pt}
  \end{subfigure}
  \begin{subfigure}[b]{0.95\linewidth}
    \includegraphics[width=\linewidth]{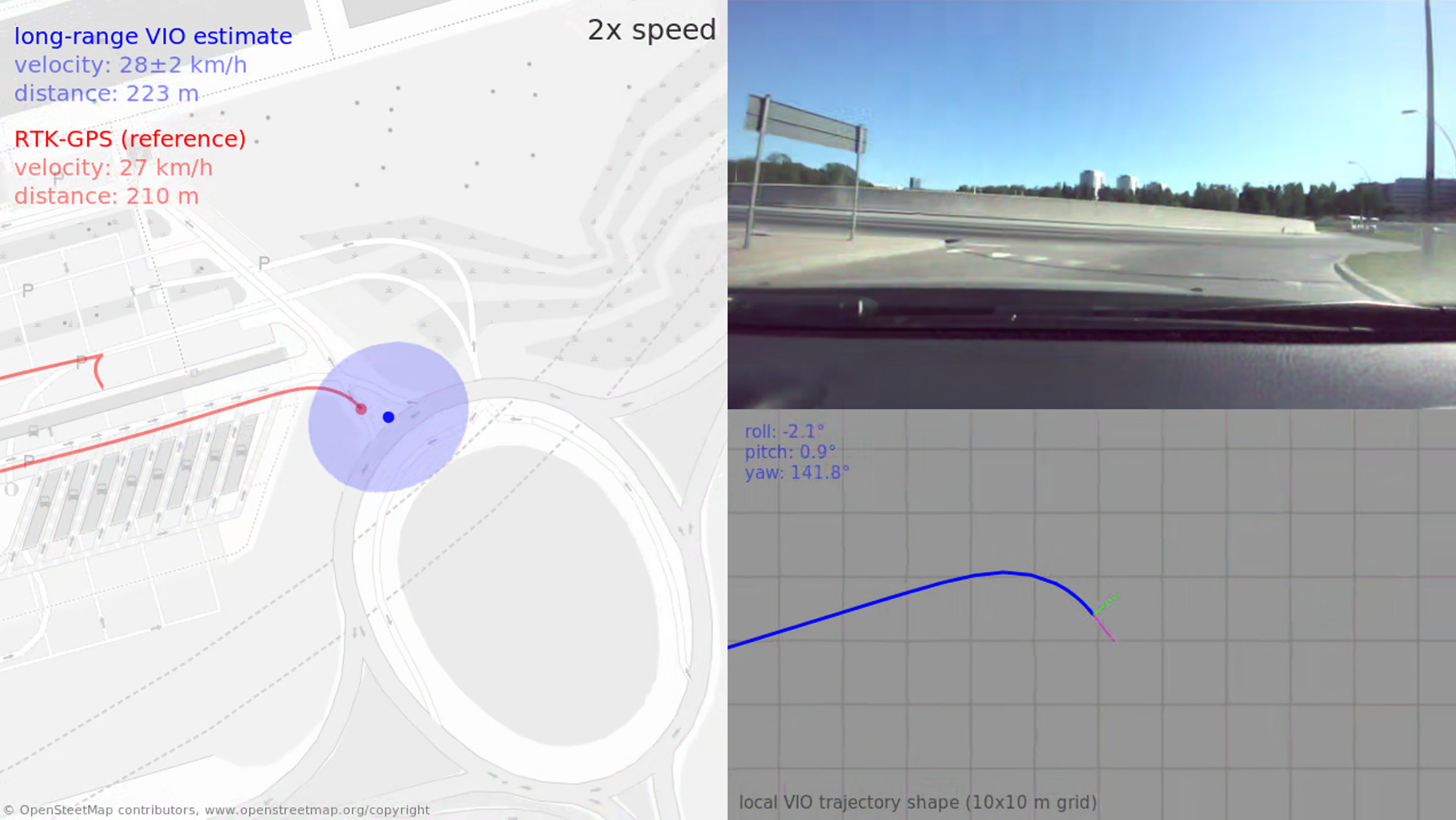}
    \caption{{\tt {vehicular-experiment-6}}}\vspace{8pt}
  \end{subfigure}
  \caption{Screenshots from the supplementary videos}
\end{figure}

\end{document}